\documentclass[mnsc,nonblindrev]{informs3_hide}
\OneAndAHalfSpacedXI 

\usepackage[english]{babel}
\usepackage[autostyle, english = american]{csquotes}
\MakeOuterQuote{"}
\usepackage[colorlinks,linkcolor=blue, citecolor=blue]{hyperref}
\usepackage[normalem]{ulem}

\usepackage{cleveref}
\crefname{subsection}{section}{subsections}
\usepackage{eqnarray}
\usepackage{algorithm,algpseudocode}
\usepackage{amsmath, bbm, enumitem, xparse, bm, lipsum}
\usepackage{amssymb}

\newcommand{\eps}{\varepsilon}

\newcommand{\Rad}{\mathrm{Rad}}

\newcommand{\VI}{V^{\mathrm{ALP}}}

\newcommand{\bc}{\bm{c}}
\newcommand{\by}{\bm{y}}

\newcommand{\bx}{\bm{x}}

\newcommand{\blue}{}
\newenvironment{blueblock}
  {\begingroup\color{black}}
  {\endgroup}

\definecolor{newrevision}{RGB}{0,0,0}
\newenvironment{newrevblock}
  {\begingroup\color{newrevision}}
  {\endgroup}

\usepackage[dvipsnames]{xcolor}

\newcommand{\newrev}[1]{\textcolor{newrevision}{#1}}

\usepackage{xparse}
\usepackage{comment}

\NewDocumentEnvironment{myproof}{o}
{\IfNoValueTF{#1}{\paragraph{{Proof.} }} {\paragraph{{#1.} }} }
{\hfill$\Halmos$}

\usepackage{natbib,bbm,multirow,multicol}
 \bibpunct[, ]{(}{)}{,}{a}{}{,}%
 \def\bibfont{\small}%
\TheoremsNumberedThrough     
\ECRepeatTheorems

\EquationsNumberedThrough    

\allowdisplaybreaks

\begin{document}


\RUNAUTHOR{Jiang, Ye, Zong}

\RUNTITLE{Resolving LP Methods for Reinforcement Learning}

\TITLE{
\Large Adaptive Resolving Methods for Markov Decision Processes with Function Approximations
}

\ARTICLEAUTHORS{%
\AUTHOR{
$\text{Jiashuo Jiang}^{\dag}$,
$\text{Yinyu Ye}^{\ddag}$,
$\text{Yiming Zong}^{\dag *}$
}

\AFF{\ \\
$\dag~$Department of Industrial Engineering \& Decision Analytics,
The Hong Kong University of Science and Technology\\
$\ddag~$Department of Management Science \& Engineering,
Stanford University
}}

\ABSTRACT{
Learning the optimal policy for Markov decision process problems (MDPs) from samples is a fundamental problem in online and data-driven decision-making. Function approximations are usually deployed to handle large or infinite state-action space. In our work, we consider the MDP problems with function approximation and we develop a new algorithm to solve it efficiently. Our algorithm is based on a linear programming (LP) reformulation and repeatedly resolves the identified reduced
linear system as new transition samples arrive. \begin{blueblock}
After the optimal basis is identified, we show that, after $N$ resolving
rounds, the expected averaged iterate achieves an instance-dependent
$\widetilde O(C_{\mathrm{inst}}/N)$ objective shortfall and signed
constraint residual. We separately account for the historical samples
used for basis identification and the $d_2$ transition queries used in
each resolving round, which yields the corresponding total
transition-query complexity.
\end{blueblock} We further complement our result with a \textit{robust} $O(1/\sqrt{N})$ bound that is independent of $\Delta$. In comparison to the guarantees established in the previous literature, our instance-dependent guarantee is tighter when the underlying instance is favorable, and the numerical experiments also reveal the wide applications and efficient empirical performances of our algorithms. 
}



\maketitle
\begingroup
\renewcommand{\thefootnote}{\fnsymbol{footnote}}
\footnotetext[1]{Corresponding author: Yiming Zong (yzongac@connect.ust.hk).}
\footnotetext[0]{Authors are listed in alphabetical order by surname.}
\footnotetext{\textcopyright~2026.
This manuscript version is made available under the
CC BY-NC-ND 4.0 license:
\url{https://creativecommons.org/licenses/by-nc-nd/4.0/}.}
\endgroup
\section{Introduction}
\label{Introduction}
Reinforcement learning (RL) plays a crucial role in navigating uncertain environments, aiming to maximize rewards by iteratively interacting with and learning from the unknown surroundings. Markov Decision Processes (MDPs) serve as a widely adopted framework for modeling environmental dynamics. They have been pivotal in diverse fields like inventory management \citep{alvo2023neural}, video gaming \citep{mnih2013playing}, robotics \citep{kober2013reinforcement}, recommender systems \citep{shani2005mdp}, and more. One of the keys to the successful deployment of RL is the use of \textit{function approximator}, which can handle large or infinite state-action space of the underlying MDPs.

In this paper, we focus on developing reinforcement learning (RL) algorithms with function approximations that are provably efficient in handling large or infinite state-action spaces. One common approach is to frame the problem as linear programming (LP) (e.g. \cite{de2003linear}), which is closely tied to approximate dynamic programming. Previous research (e.g. \cite{ozdaglar2023revisiting}) has established a worst-case $O(1/\sqrt{N})$ suboptimality gap, which translates into $O(1/\epsilon^2)$ sample complexity for this LP-based method, representing the best possible guarantee achievable for the most difficult problems in this problem class.
However, these minimax bounds and worst-case analyses can often be overly cautious, leading to a gap between theoretical assurances and practical performance in specific instances. We focus on a more tailored approach - one that guarantees outstanding performance for each unique problem and offers problem-specific assurances. Our study progresses by introducing RL algorithms that provide problem-specific guarantees.



\subsection{Main Results and Contributions}

We now summarize our main results and contributions. Our main result is to propose a new RL algorithm with function approximations that enjoy a problem/instance-dependent theoretical guarantee. Our algorithm is LP-based, which is built upon previous LP-based methods for approximate dynamic programming (e.g. \cite{de2003linear}) or RL with linear function approximation (e.g. \cite{ozdaglar2023revisiting}). However, our algorithm differs from previous ones and we make the following innovations.

\textbf{Constraints reduction and variables deletion}.
We first show that the LP size can be significantly reduced even when the state-action space is large or infinite. Existing LP formulations for RL, including approximate linear
programming (ALP) and reduced linear programming (RLP)
\cite{puterman1994markov,de2003linear,de2004constraint},
can still contain a very large number of constraints, making
their direct data-driven solution sample inefficient. We prove that both the number of constraints and decision variables can be reduced to at most the number of basis functions, independent of the state-action space size. Based on this observation, we develop a data-driven basis identification algorithm (\Cref{alg:Idenbasis}) and prove that the reduced LP preserves the optimal solution with high probability.

\textbf{Learn the optimal weights}.
After identifying the reduced LP, we adopt the resolving framework to learn the optimal weights. Unlike existing resolving methods, \Cref{alg:Twophase} resolves the induced linear system rather than the LP itself, further improving sample efficiency.
\begin{blueblock}
Conditional on successful basis identification, the expected averaged
iterate of \Cref{alg:Twophase} satisfies an instance-dependent
$\widetilde O(C_{\mathrm{inst}}/N)$ objective and signed-feasibility
guarantee after $N$ resolving rounds. Combining this resolving-phase
bound with the samples required for basis identification gives an
instance-dependent $\widetilde O(1/\epsilon)$ dependence on the target
accuracy in the total transition-query complexity.
\end{blueblock}

\textbf{Comparison to \cite{jiang2024achieving}}. Note that the resolving algorithms have been developed for constrained MDP problem in \cite{jiang2024achieving} under the tabular setting, however, our algorithms differ from theirs in several aspects. First, in our problem, the state-action space can be infinite or continuous, while the state-action space in \cite{jiang2024achieving} is finite. Second, our algorithm to identify the optimal basis only requires solving the LP once by using the simplex method, while the basis identification algorithm in \cite{jiang2024achieving} requires solving the LP multiple times. Our algorithm is more computationally efficient and is more suitable to large-scale problems. Finally, our algorithm enjoys a robust $O(1/\epsilon^2)$ guarantee while \cite{jiang2024achieving} establishes only the instance-dependent $\tilde{O}(1/\epsilon)$ guarantee under the tabular setting.




\subsection{Related Works}

Our paper is further related to two lines of work: i) instance-dependent bound for RL, and ii) near-optimal algorithms for online resource allocation. We now briefly review the two lines of research below.

\textbf{Instance-dependent bound for RL}.
There has been growing interest in instance-dependent guarantees for RL under different settings. Existing works have established instance-dependent regret bounds for linear and linear mixture MDPs \cite{he2021logarithmic, kim2021improved, zhou2022computationally}, with subsequent extensions to heavy-tailed rewards \cite{li2023variance, huang2024tackling}. In the tabular setting, \citet{wagenmaker2022beyond} showed that instance-dependent sample complexity can significantly improve upon worst-case guarantees, while \citet{al2023towards} established the first instance-dependent lower bound and the PEDEL algorithm \cite{wagenmaker2022instance} nearly matches this lower bound. Our work differs from the above literature by developing a new LP-based resolving algorithm for RL with general function approximation and establishing an instance-dependent sample complexity guarantee.

\textbf{Near-optimal algorithms for online resource allocation}. 
Online resource allocation has been extensively studied under LP formulations, with applications including online knapsack, network revenue management, matching, and other online allocation problems \citep{ gallego1997multiproduct, mehta2007adwords, molinaro2014geometry, agrawal2014dynamic}. Existing works mainly consider two input models: the stochastic input model and the random permutation model \citep{agrawal2014dynamic, kesselheim2014primal, gupta2014experts}. Under a non-degeneracy assumption, logarithmic regret bounds have been established for network revenue management, general online LP, and convex allocation problems \citep{jasin2014reoptimization, li2022online, ma2022optimal}. More recent works have further removed or relaxed this assumption, leading to logarithmic or even constant regret guarantees under broader settings \citep{bumpensanti2020re, jiang2022degeneracy, ao2025learning, banerjee2024good}.

\section{Preliminaries}\label{sec:Infinitedis}
We consider a MDP problem with the state space denoted by $\mathcal{S}$ and the action space denoted by $\mathcal{A}$. We denote by $\gamma\in(0,1)$ a discount factor. We also denote by $P: \mathcal{S}\times\mathcal{A}\rightarrow \mathcal{D}(\mathcal{S})$ the probability transition kernel of the MDP, where $\mathcal{D}(\mathcal{S})$ denotes a probability measure over the state space $\mathcal{S}$. Then, $P(s'|s, a)$ denotes the probability of transiting from state $s\in\mathcal{S}$ to state $s'\in\mathcal{S}$ when the action $a\in\mathcal{A}$ is executed. The initial distribution over the states of the MDP is denoted by $\mu_1$.

There is a cost function \blue{$c: \mathcal{S}\times\mathcal{A}\rightarrow[0,1]$}. 
We focus on the Markovian policy, which takes the action only based on the current state of the MDP. To be specific, any Markovian policy $\pi$ can be denoted as a function $\pi: \mathcal{S}\rightarrow \mathcal{A}$.
For any Markovian policy $\pi$, we denote by $V^\pi(\mu_1)$ the infinite horizon discounted cost of the policy $\pi$, with the formulation of $V^{\pi}(\mu_1)$ given below:
\begin{equation}\label{eqn:Disreward}
V^{\pi}(\mu_1)=\mathbb{E}\left[ \sum_{t=1}^{\infty}\gamma^{t-1}\cdot c(s_t, a_t)\mid \mu_1 \right],
\end{equation}
where $(s_t, a_t)$ is generated according to the policy $\pi$ and the transition kernel $P$ with the initial state distribution $\mu_1$.
To solve the MDP problem, we aim to find an optimal Markovian policy, denoted by $\pi^*$, that minimizes the cost in \eqref{eqn:Disreward}. Importantly, we assume that the transition kernel $P$ is \textit{unknown} to the decision maker. We obtain samples to learn the transition kernel. The sampling procedure can be described as follows.

\begin{assumption}[Generative Model]\label{assump:1}
For each state and action pair $(s,a)$, we can query the model $\mathcal{M}$ to obtain an observation of the new state $s'\in\mathcal{S}$, where the transition from $s$ to $s'$ follows the probability kernel $P(s'|s,a)$ independently.
\end{assumption}


Querying the generative model $\mathcal{M}$ can be costly, and it is desirable to approximate the optimal policy $\pi^*$ well with as few samples as possible. Therefore, we measure the performance of a policy $\pi$ by the \textit{sample complexity} bound. That is, for any $\eps$, we compute a bound on the number of samples that we need to construct a policy $\pi$ such that
\begin{equation}\label{eqn:complexity}
V^{\pi}(\mu_1) - V^{\pi^*}(\mu_1) \leq\eps.
\end{equation}
\begin{blueblock}
Our main analysis first controls the LP objective shortfall and signed
constraint residual of the expected averaged iterate. The translation
from this LP guarantee to a policy-performance guarantee requires the
additional full-ALP, exact-realizability, and concentrability conditions
stated in \Cref{cor:value_function_policy}.
\end{blueblock}
The Bellman optimality equation can be written as
\begin{equation}\label{eqn:Bellman}
\begin{aligned}
&V^{\pi^*}(s)=\mathop{\min}\limits_{a\in\mathcal{A}}c(s,a)+\gamma\sum_{s'\in\mathcal{S}}P(s'|s,a)V^{\pi^*}(s').
\end{aligned}
\end{equation}
Note that \eqref{eqn:Bellman} implies that the optimal policy $\pi^*$ is the argmin-greedy policy with respect to 
the optimal value function $V^{\pi^*}$ through the Bellman equation \eqref{eqn:Bellman}. Thus, in order to approximate the optimal policy $\pi^*$, it is sufficient to approximate 
the value function $V^{\pi^*}$.

\subsection{Approximate Dynamic Programming and Linear Programming}\label{sec:ADP}

One can directly compute the optimal policy $\pi^*$ by solving the dynamic programming induced by \eqref{eqn:Disreward}. However, when the state space $\mathcal{S}$ or the action space $\mathcal{A}$ is very large or infinite, solving the DP becomes computationally intractable. Various methods have been developed in the literature to tackle this issue (e.g. \cite{powell2007approximate, li2006towards, sutton2018reinforcement}), and we consider the function approximations. To be specific, we make the following assumption
\begin{assumption}\label{assump:FunctionApp}
The value function $V^{\pi^*}$ can be well approximated by
 \begin{equation}\label{eqn:Approx}
    V^{\pi^*}(s)\approx \sum_{i=1}^{d_1} \phi_i(s)\cdot w_i, ~~\forall s\in\mathcal{S},
\end{equation}
where $\phi_1, \dots, \phi_{d_1}$ are some basis functions that are determined based on the particular problem instances, and $\bm{w}=(w_1,\dots, w_{d_1})\in\mathbb{R}^{d_1}$ are the weights to be determined. \blue{Moreover, the basis functions are normalized so that
\[
0\leq \phi_i(s)\leq 1,
\qquad
s\in\mathcal S,\quad i\in[d_1].
\]}
\end{assumption}
Then, the approximate linear programming (ALP) has been developed in the previous literature (e.g. \cite{puterman1994markov}) to find the optimal weights $\bm{w}$ that well approximates the optimal value function. The ALP is formulated as follows where the weights $\bm{w}$ are regarded as the decision variables.
\begin{subequations} \label{lp:Approx}
\begin{align}
V^{\mathrm{ALP}}=~\max\ & \sum_{s\in\mathcal{S}} \mu(s)\cdot \sum_{i=1}^{d_1}\phi_i(s)w_i \label{eqn:Approxobj}
\\ \mathrm{s.t.\ }& 
\sum_{i=1}^{d_1}\phi_i(s)w_i - \gamma\cdot \sum_{s'\in\mathcal{S}} P(s'|s,a)\label{eqn:Approxcon}
\\&\cdot \sum_{i=1}^{d_1}\phi_i(s')w_i
\leq c(s,a), ~~\forall (s,a)\in\mathcal{S}\times\mathcal{A} \nonumber
\\ & \bm{w}\in\mathbb{R}^{d_1}.  \label{eqn:Approxconstraint3}
\end{align}
\end{subequations}
where $\mu$ is a \textit{state-relevance weights} with positive elements. 
The ALP has been widely studied in the literature and numerous methods have been developed to solve it efficiently. In what follows, we focus on solving the ALP \eqref{lp:Approx} in a data-driven way, where we aim to obtain a near-optimal solution with as few samples as possible.

\subsection{Reduced LP for Large or Infinite State Space}\label{sec:extension}

When the underlying state-action space $\mathcal{S}\times\mathcal{A}$ is large or infinite, the ALP \eqref{lp:Approx} will have a large number of or infinite constraints and thus intractable to solve. A common approach to deal with this issue in the literature is through constraint sampling.
To be specific, following \cite{reemtsen2001semi, de2004constraint}, we can sample a finite subset $\mathcal{K}\subset\mathcal{S}\times\mathcal{A}$, and we consider the reduced LP (RLP) given as follows,
\begin{subequations} \label{lp:Reduced}
\begin{align}
V^{\mathrm{RLP}}=~\max\ & \sum_{s\in\mathcal{S}} \mu(s)\cdot \sum_{i=1}^{d_1}\phi_i(s)w_i \label{eqn:Reobj}
\\ \mathrm{s.t.\ }& \sum_{i=1}^{d_1}\phi_i(s)w_i - \gamma\cdot \sum_{s'\in\mathcal{S}} P(s'|s,a) \label{eqn:Recon}\\&\cdot \sum_{i=1}^{d_1}\phi_i(s')w_i \leq c(s,a), ~~\forall (s,a)\in\mathcal{K} \nonumber
\\ & \bm{w}\in\mathbb{R}^{d_1}.  \label{eqn:Reconstraint3}
\end{align}
\end{subequations}
We have the following result regarding the size of $V^{\text{RLP}}$.
\begin{theorem}\label{thm:ConstSample}
(Theorem 3.1 of \cite{de2004constraint}) Let the elements in the set $\mathcal{K}$ be sampled independently from $\mathcal{S}\times\mathcal{A}$. Then, for any $\epsilon, \delta>0$, when $|\mathcal{K}| = O\left(\frac{\log(1/\epsilon)}{\epsilon}\cdot\log(1/\delta)\right)$, it holds that $P\left( |\VI-V^{\mathrm{RLP}}|\leq\eps\right)\geq1-\delta$.
\end{theorem}
Although the dimension of $V^{\mathrm{RLP}}$ is finite, it can still be very large which requires a large amount of data to be solved. We then develop an approach that could further reduce the size of the LP. 

\section{Variables and Constraints Reduction}
Though we can use samples to construct an estimation of ALP \eqref{lp:Approx} (or RLP \eqref{lp:Reduced} for large problems) and solve it directly, the dimension of the LP can be large, which implies that it may require many samples to approximate the optimal solution accurately. Thus, directly solving the LP is not sample efficient. 
In this section, we describe a procedure that can further reduce the required number of constraints to a quantity $d_2$ that is guaranteed to be no larger than the number of basis functions $d_1$. We develop a high probability bound showing that the optimal solution can still be effectively found with finite samples. 
For convenience, we denote by the ALP (or RLP for large problems) in the following standard matrix form. We set $K$ to be the number of constraints in the LP, with $K=|\mathcal{S}\times\mathcal{A}|$ for a finite ALP and $K=|\mathcal{K}|$ for an RLP.
\begin{blueblock}
Throughout Sections~3--5, the finite-sample analysis is applied to a finite LP. When the original state-action space is infinite, we first construct the finite RLP in \eqref{lp:Reduced}, and all subsequent
union bounds and basis arguments are applied to that finite LP.
\end{blueblock}
\begin{equation} \label{lp:matrix}
V^{\mathrm{LP}}=~\max~ \bm{r}^\top \bm{x} ~~~ \mathrm{s.t.} ~A\bm{x} +\bm{s} = \bm{c},~~~ \bm{x}\in\mathbb{R}^{d_1}, \bm{s}\geq0.
\end{equation}
Here, we use $\bm{r}=\bm{\mu}^\top\Phi$, and the matrix $A\in\mathbb{R}^{K\times d_1}$ denotes the constraint matrix, with \blue{$A_{(s,a),i}:=\phi_i(s)-\gamma\mathbb{E}[\phi_i(s')\mid s,a]$}. We set $\bm{s}\in\mathbb{R}^K$ to be the slackness variables. We now show that when solving \eqref{lp:matrix}, instead of focusing on all the constraints (the number of which may be large), we can focus on only a small set of constraints that is no larger than the number of decision variables. \blue{We impose the following local regularity condition on the finite LP and on the deterministic basis-selection rule used by Algorithm~\ref{alg:Idenbasis}. This condition is used in the local perturbation analysis of Theorems~\ref{thm:Infibasis2} and~\ref{thm:Robust}.}
\begin{assumption}
\label{assump:degeneracy}
The population LP $V^{\mathrm{LP}}$ is nondegenerate and has a unique optimal basis. \blue{There exist $\eta_{\mathrm{reg}}>0$ and $B_x<\infty$ such that, for every $\widetilde A$ with $\|\widetilde A-A\|_{\max}\leq\eta_{\mathrm{reg}}$, the prescribed rule applied to $\max_{\bm x}\{\bm r^\top\bm x: \widetilde A\bm x\leq\bm c\}$ returns an optimal basic solution $\widetilde{\bm x}$ and basis $(\widetilde{\mathcal I},\widetilde{\mathcal J})$ satisfying $\|\widetilde{\bm x}\|_1\leq B_x$ and $A_{\widetilde{\mathcal J}, \widetilde{\mathcal I}}$ nonsingular.}
\end{assumption}

\subsection{Optimal Basis Identification}
Following standard LP theory, when solving the LP \eqref{lp:matrix}, it is sufficient for us to focus only on the \textit{corner points} of the feasible set to obtain an optimal solution. Such a solution is also called \textit{basic solutions}. Therefore, we can simply focus on obtaining the optimal basic solution of LP \eqref{lp:matrix}.
The optimal basic solution of LP \eqref{lp:matrix} enjoys the following further characterizations. Note that in LP theory, the corner point can be represented by \textit{LP basis}, which involves the set of basic variables that are allowed to be non-zero, and the set of active constraints that are binding under the corresponding basic solution.
Then, we have the following result, following standard LP theory.

\begin{lemma}\label{lem:Basis}
There exists an index set $\mathcal{I}^*\subset [d_1]$ and an index set $\mathcal{J}^*\subset \mathcal{S}\times\mathcal{A}$ such that $|\mathcal{I}^*|=|\mathcal{J}^*|=d_2$, for some integer $d_2\leq d_1$. Also, for the given $\mathcal{I}^*$ and $\mathcal{J}^*$, there exists an optimal solution $\bm{x}^*$ to LP \eqref{lp:matrix} that satisfies
\begin{equation}\label{eqn:Lsystem}
    A_{\mathcal{J}^*, \mathcal{I}^*} \cdot\bm{x}^*_{\mathcal{I}^*} = \bm{c}_{\mathcal{J}^*}
\end{equation}
and $\bm{x}^*_{\mathcal{I}^{*c}}=0$ where $\mathcal{I}^{*c}=[d_1]\setminus\mathcal{I}^*$ denotes the complementary set of $\mathcal{I}^*$.
\end{lemma}
Therefore, characterizing the optimal corner point is equivalent to identifying the optimal index sets $\mathcal{I}^*$ and $\mathcal{J}^*$. We now describe how to identify one optimal basis of LP \eqref{lp:matrix}. Note that since the transition kernel $P$ is unknown, we can only use the historical dataset, denoted by $\mathcal{H}$, to construct an estimate of $A$, which we denote by $\hat{A}(\mathcal{H})$. We describe in the following paragraphs the exact procedure to construct estimate $\hat{A}(\mathcal{H})$. With $\hat{A}(\mathcal{H})$, we can consider the following LP, which serves as an estimate of LP \eqref{lp:matrix}.
\begin{equation} \label{lp:estimate}
\begin{aligned}
\hat{V}^{\mathrm{LP}}(\mathcal{H})=\max~ &\bm{r}^\top \bm{x} \\
\mathrm{s.t.} ~~~&\hat{A}(\mathcal{H})\cdot\bm{x} +\bm{s} = \bm{c},\\
&\bm{x}\in\mathbb{R}^{d_1}, \bm{s}\geq0.
\end{aligned}
\end{equation}
A key step in our approach is to identify one optimal basis of $\hat{V}^{\mathrm{LP}}(\mathcal{H})$ and then show that it is a good approximation of the optimal basis of $V^{\mathrm{LP}}$. 

\blue{Optimal-basis identification is a standard output of the simplex method. Given the empirical matrix $\widehat A(\mathcal H)$, Algorithm~\ref{alg:Idenbasis} solves the empirical LP once and extracts its basic weight variables and nonbasic slack variables.}

\begin{algorithm}[t] 
\caption{Optimal-basis identification} \label{alg:Idenbasis} 
\begin{algorithmic}[1] 
\State \blue{\textbf{Input:} Historical transition dataset $\mathcal H$.}
\State \blue{Construct the empirical LP $\widehat V^{\mathrm{LP}}(\mathcal H)$ in~\eqref{lp:estimate} and solve it by the simplex method. Let $\widehat{\bm x}$ and $\mathcal B^N$ denote the optimal basic solution and basis selected by the fixed deterministic tie-breaking rule.}
\State \blue{Let $\mathcal I^N$ be the indices of the basic weight variables and let $\mathcal J^N$ be the indices of the nonbasic slack variables.} 
\State \blue{\textbf{Output:} $\widehat{\bm x}$, $\mathcal I^N$, and $\mathcal J^N$.}
\end{algorithmic} 
\end{algorithm}

\blue{Let $\widehat d_2:=|\mathcal I^N|=|\mathcal J^N|.$ Standard basis structure gives $\widehat d_2\leq d_1$ and implies that $\widehat A_{\mathcal J^N,\mathcal I^N}(\mathcal H)$ is square and nonsingular. Thus, $\mathcal I^N$ and $\mathcal J^N$ define the reduced empirical system used in Algorithm~\ref{alg:Twophase}. The corresponding LP argument is given in Appendix~\ref{app:D}.}

\subsection{High Probability Bound of Optimality and Finite Sample Guarantee}

We know that \Cref{alg:Idenbasis} identifies an optimal basis of $\hat{V}^{\mathrm{LP}}(\mathcal{H})$.
In this section, we further show that the output of \Cref{alg:Idenbasis}, denoted by $\mathcal{I}^N$ and $\mathcal{J}^N$, forms an optimal basis to $V^{\mathrm{LP}}$ with a high probability, where the randomness comes from the randomness of the dataset $\mathcal{H}$ that we use to construct estimate of $A$.

There are two key quantities that we need to specify in order to derive the high probability bound. For an arbitrary basis $\mathcal{I}$ and $\mathcal{J}$ to $V^{\mathrm{LP}}$, satisfying $|\mathcal{I}|=|\mathcal{J}|$, a solution $\bx(\mathcal{I}, \mathcal{J})$ satisfying the characterizations in \Cref{lem:Basis} is called a \textit{basic solution} if $A_{\mathcal{J}, \mathcal{I}}$ is a non-singular sub-matrix such that the solution to linear equations \eqref{eqn:Lsystem} is uniquely determined. We now denote by $\mathcal{F}_1$ the collection of all basis $(\mathcal{I}, \mathcal{J})$ such that $\bx(\mathcal{I}, \mathcal{J})$ is a basis solution. Note that not all basic solutions will be feasible. Therefore, we define $\delta_1$ as the \textit{feasibility gap}, specified below.
\begin{equation*}\label{def:FeasiGap}
\begin{aligned}
\delta_1 =&\min_{(\mathcal{I}, \mathcal{J})\in\mathcal{F}_1}\Biggl\{ \max_{k\in[\blue{K}]}\left\{\left[  A_{k, :}\cdot\bx(\mathcal{I}, \mathcal{J}) - c_k \right]^+ \right\}: \\
&~~\max_{k\in[\blue{K}]}\left\{\left[  A_{k, :}\cdot\bx(\mathcal{I}, \mathcal{J}) - c_k \right]^+ \right\} > 0 \Biggr\}
\end{aligned}
\end{equation*}
Note that if a basic solution $\bx(\mathcal{I}, \mathcal{J})$ is feasible, then 
$$\max_{k\in[\blue{K}]}\left\{\left[  A_{k, :}\cdot\bx(\mathcal{I}, \mathcal{J}) - c_k \right]^+ \right\} = 0.$$ However, if $\bx(\mathcal{I}, \mathcal{J})$ is infeasible, then there must exist a $k\in[K]$ such that $\left[  A_{k, :}\cdot\bx(\mathcal{I}, \mathcal{J}) - c_k \right]^+ > 0$ becomes an infeasibility gap over the constraints. The parameter $\delta_1$ specifies the minimum of such a gap. Since the number of basis is always finite, we know that it must hold $\delta_1>0$ (the situation where all basic solutions are feasible is discussed later). 

We also need a parameter that characterizes of the suboptimality gap of feasible basic solutions. Denote by $\mathcal{F}_2$ the collection of all feasible basis such that $\bx(\mathcal{I}, \mathcal{J})$ is feasible to $V^{\mathrm{LP}}$ for all $(\mathcal{I}, \mathcal{J}) \in \mathcal{F}_2$. Then, we define 
\begin{equation*}\label{def:OptGap}
    \delta_2 = \min_{(\mathcal{I}, \mathcal{J})\in\mathcal{F}_2:V^{\mathrm{LP}} > \bm{r}^\top\bx(\mathcal{I}, \mathcal{J})}\left\{ V^{\mathrm{LP}} - \bm{r}^\top\bx(\mathcal{I}, \mathcal{J})\right\}. 
\end{equation*}
Briefly speaking, $\delta_2$ specifies the minimum suboptimality gap between the optimal solution and the best sub-optimal basic solution of $V^{\mathrm{LP}}$. We now define
\begin{equation}\label{def:delta}
    \Delta = \min\{\delta_1, \delta_2\}.
\end{equation}
We adopt the convention that the minimum over an empty set is $+\infty$.
\blue{If exactly one of the two minimization domains is empty, then \(\Delta\) equals the other finite gap. For all \(\Delta\)-dependent results below, we assume that the two minimization domains are not simultaneously empty. Thus, the special case where every basic solution is feasible and optimal is excluded from the \(\Delta\)-dependent results, while the \(\Delta\)-independent robust guarantee remains applicable.} The parameter $\Delta$ specifies distance between an optimal basis solution and other non-optimal or non-feasible basic solutions.

One crucial part of our analysis is to show that, when \blue{the entrywise estimation error is no larger than the effective basis-identification margin introduced below}, the output of \Cref{alg:Idenbasis} is an optimal basis of $V^{\mathrm{LP}}$. To \blue{quantify this estimation error}, we define the following quantity:
\begin{equation}\label{eqn:Rad}
\Rad(N,\blue{\eps_0})=\sqrt{\frac{\log(2/\blue{\eps_0})}{2N}}.
\end{equation}
Suppose that the dataset $\mathcal{H}$ contains $N$ transition data of the state-action pair $(s,a)$. Denote by $\{s_1, \dots, s_N\}$ the state transition. Then, the element of $A$ at the row $(s,a)$ and column $i$ can be \blue{estimated by $$\hat A_{(s,a),i}(\mathcal H)
:=\phi_i(s) - \frac{\gamma}{N}\cdot\sum_{n=1}^N \phi_i(s_n).$$} Following the standard Hoeffding's inequality, we know that the gap between \blue{$\hat A_{(s,a),i}(\mathcal H)$} and $A_{(s,a), i}$ is upper bounded by $\Rad(N, \blue{\eps_0})$ with probability at least \blue{$1-\eps_0$}. We now present the theorem showing that \Cref{alg:Idenbasis} indeed helps us identify one optimal basis with a high probability.

\blue{The basis-identification argument involves both the separation gap $\Delta$ and local perturbation-stability constants. Let $\eta_{\mathrm{id}}>0$ and $C_{\mathrm{id}}<\infty$ denote the instance-dependent constants specified in Appendix~\ref{app:E}, and define the effective basis-identification margin
\begin{equation}
\label{def:DeltaId}
\Delta_{\mathrm{id}}
:=
\min\left\{
\eta_{\mathrm{id}},
\frac{\Delta}{2C_{\mathrm{id}}}
\right\}.
\end{equation}
The constant $\eta_{\mathrm{id}}$ specifies the neighborhood in which the relevant empirical and population basis systems are stable, while $C_{\mathrm{id}}$ converts entrywise matrix estimation error into feasibility and objective perturbations. Both constants are independent of the sample size and the confidence level.}
\begin{blueblock}
\begin{theorem}
\label{thm:Infibasis2}
Under Assumption~\ref{assump:degeneracy}, let
$\epsilon>0$ and set $\epsilon_0=\frac{\epsilon}{d_1K}.$
If Algorithm~\ref{alg:Idenbasis} uses $N_{\mathrm I}$
historical transitions per constraint and $\operatorname{Rad} \left( N_{\mathrm I},\epsilon_0 \right) \leq \Delta_{\mathrm{id}},$ then
\[
\mathbb P
\left(
\mathcal I^{N_{\mathrm I}}=\mathcal I^*,
\;
\mathcal J^{N_{\mathrm I}}=\mathcal J^*
\right)
\geq
1-\epsilon.
\]
\end{theorem}

In particular, it is sufficient to take $N_{\mathrm I}\geq\left\lceil\frac{\log(2d_1K/\epsilon)}{2\Delta_{\mathrm{id}}^2}\right\rceil,$ and hence Algorithm~\ref{alg:Idenbasis} uses $O\left(\frac{K\log(2d_1K/\epsilon)}{\Delta_{\mathrm{id}}^2}\right)$ historical transition queries. When the gap-dependent term is active,
the historical query bound has the usual
$O(\Delta^{-2})$ dependence. The following $\Delta$-independent guarantee holds for any
fixed sample size.

\end{blueblock}

\begin{blueblock}
\begin{theorem}
\label{thm:Robust}
Under Assumption~\ref{assump:degeneracy}, let
$\epsilon>0$ and set $\epsilon_0 = \frac{\epsilon}{d_1K}.$ If $\operatorname{Rad}(N,\epsilon_0) \leq \eta_{\mathrm{id}},$ then, with probability at least $1-\epsilon$, the output $\widehat{\bm x}$ of Algorithm~\ref{alg:Idenbasis} satisfies
\[
\left|
V^{\mathrm{LP}}
-
\bm r^\top\widehat{\bm x}
\right|
\leq
C_{\mathrm{val}}
\operatorname{Rad}(N,\epsilon_0),
~~
A\widehat{\bm x}
\leq
\bm c
+
B_x
\operatorname{Rad}(N,\epsilon_0)\bm1.
\]
\end{theorem}
\end{blueblock}

\section{Our Formal Algorithm}
In the previous section, we described how to identify one optimal basis $\mathcal{I}^*$ and $\mathcal{J}^*$. We now describe how to approximate the optimal weights $\bm{x}^*$ that corresponds to the optimal basis $\mathcal{I}^*$ and $\mathcal{J}^*$. Note that the optimal solution $\bm{x}^*$ enjoys the following structure: $\bm{x}^*_{\mathcal{I}^{*c}}=0$ and the non-zero elements $\bm{x}^*_{\mathcal{I}^*}$ can be given as the solution to
\begin{equation}\label{eqn:OptQ}
A_{\mathcal{J}^*, \mathcal{I}^*}\cdot \bm{x}_{\mathcal{I}^*}=\bm{c}_{\mathcal{J}^*}.
\end{equation}
However, in practice, we do not know the optimal basis $\mathcal{I}^*$ and $\mathcal{J}^*$, as well as the matrix $A$, beforehand. Therefore, on one hand, we use \Cref{alg:Idenbasis} to learn the optimal basis $\mathcal{I}^*$ and $\mathcal{J}^*$. On the other hand, we construct estimates of the matrix $A$ using the historical dataset. \blue{We denote by $C$ a known constant satisfying $C\geq2\|\bx^*\|_1$.} Our formal algorithm is given in \Cref{alg:Twophase}.
\begin{blueblock}
Here $M^\dagger$ denotes the Moore--Penrose pseudoinverse and Lemma~\ref{lem:projection} shows that it equals $M^{-1}$ on the event used in the resolving analysis. For $k=(s,a)\in\mathcal J^N$ and $i\in\mathcal I^N$, $A^n_{k,i}=\phi_i(s)-\gamma\phi_i(s'(s,a))$ is conditionally unbiased for $A_{k,i}$. The adaptive remaining-capacity process is analyzed in Appendix~\ref{pf:Thm2}. The output $\bar x^N$ of \Cref{alg:Twophase} is random, and we define its mean output by $x^N_{\text{mean}}:=\mathbb E[\bar x^N] = \frac{1}{N}\sum_{n=1}^N\mathbb E[x^n]$.
\end{blueblock}

Note that in our algorithm, for each iteration $n\in\blue{[N]}$, we need to query the generative model $\mathcal{M}$ to obtain an unbiased estimator of the matrix $A$ to represent how the constraints are satisfied under the current action $\bm{x}^n$. Such an estimator can be constructed in the following way. For any state-action pair $(s,a)$, we query the generative model $\mathcal{M}$ to obtain a state transition, where we denote the new state by $s'$. Then,
the element of $A^n$, at the row $(s,a)\in\mathcal{J}^N$ and column $i\in\blue{\mathcal{I}^N}$ can be represented by $\phi_i(s) - \gamma\cdot \phi_i(s')$. It is clear to see that
\begin{equation*}
\begin{aligned}
&\mathbb{E}_{s'}[A^n_{(s,a), i}] = \mathbb{E}_{s'}[\phi_i(s) - \gamma\cdot \phi_i(s')] = \\
&\phi_i(s) - \gamma\cdot\sum_{s'\in\mathcal{S}} P(s'|s,a)\cdot\phi_i(s') = A_{(s,a), i}.
\end{aligned}
\end{equation*}

\begin{algorithm}[ht!]
\caption{The Algorithm for Optimal Weights}
\label{alg:Twophase}
\begin{algorithmic}[1]
\State \textbf{Input:} \blue{the number of resolving rounds $N$, the identified basis $(\mathcal I^N,\mathcal J^N)$ returned by Algorithm~\ref{alg:Idenbasis}, and a constant $C\geq2\|\bm x^*\|_1$.}
\State \blue{Initialize $\mathcal H^1=(\mathcal H_k^1)_{k\in\mathcal J^N}$ with one fresh transition for each $k\in\mathcal J^N$}, and \blue{set} $\bm c_{\mathcal J^N}^1=N\bm c_{\mathcal J^N}$.
\For{$n=1,\dots, N$}
\State Construct estimates $\hat{A}_{\blue{\mathcal{J}^N,\mathcal{I}^N}}(\mathcal{H}^n)$ using dataset $\mathcal{H}^n$.
\State \blue{Define $\widetilde{\bm x}^n$ by}
\begin{equation}
\label{eqn:OptQ2}
\blue{\widetilde{\bm x}_{\mathcal I^N}^n = \widehat A_{\mathcal J^N,\mathcal I^N} (\mathcal H^n)^\dagger
\frac{\bm c_{\mathcal J^N}^n}{N-n+1},~\widetilde{\bm x}_{(\mathcal I^N)^c}^n=\bm 0.}
\end{equation}
\State \blue{Project $\tilde{\bm{x}}^n$ onto the $\ell_1$-ball $\{\bm{x}: \|\bm{x}\|_1\leq C\}$ to obtain $\bm{x}^n$.}
\State For each $(s,a)\in\blue{\mathcal{J}^N}$, query the generative model $\mathcal{M}$ to obtain the new state transition $s'(s,a)$.
\State \blue{For each $k=(s,a)\in\mathcal J^N$, update $\mathcal H_k^{n+1}=\mathcal H_k^n\cup\{s'(s,a)\}$.}
\blue{\State Construct
$A^n\in\mathbb R^{\widehat d_2\times\widehat d_2}$.
For $k=(s,a)\in\mathcal J^N$ and
$i\in\mathcal I^N$, set $A^n_{k,i}
=
\phi_i(s)
-
\gamma
\phi_i\!\left(s'(s,a)\right).$}
\State Do the update:
\begin{equation}\label{eqn:UpdateAlpha2}
\bm{c}^{n+1}_{\mathcal{J}^N}=\bm{c}^{n}_{\mathcal{J}^N}-A^n\cdot \bm{x}^n_{\mathcal{I}^N}.
\end{equation}
\EndFor
\State Define
\begin{equation}\label{eqn:122402}
    \bar{\bx}^N = \frac{1}{N}\cdot\sum_{n=1}^N \bx^n.
\end{equation}
\State \textbf{Output:} $\bar{\bm{x}}^N$.
\end{algorithmic}
\end{algorithm}

Another crucial element in \Cref{alg:Twophase} (step 10) is that we adaptively update the value of $\bm{c}^{n}$ as in \eqref{eqn:UpdateAlpha2}. We then use the updated $\bm{c}^{n}$ to obtain the value of $\tilde{\bm{x}}^n_{\blue{\mathcal{I}^N}}$ as in \eqref{eqn:OptQ2}. Such a resolving algorithmic design has been developed in \cite{jiang2024achieving} for constrained MDP under the tabular setting and we further develop here for RL under the linear approximation setting. The resolving procedure has a natural interpretation that \blue{the normalized remaining capacity $\bm{c}^{n} / (N-n+1)$} will concentrate around $\bm{c}$, with a gap bounded at the order of $1/(N-n+1)$, which is the key to achieving an instance-dependent guarantee. To be specific, \blue{for $n\in[N-1]$}, if for one binding constraint $j\in\blue{\mathcal{J}^N}$, the constraint value $\blue{\sum_{t=1}^n A^t_{j,:}\bx^t_{\mathcal I^N}/ {n}}$ is below the target $\bm{c}_j$, we know that $\blue{c_j^{n+1}/(N-n)}$ is greater than $\bm{c}_j$, which results in a greater constraint value for $\tilde{\bm{x}}^{n+1}$. In this way, any gap between the real-time constraint value and the target will be self-corrected in the next period as we adaptively obtain $\tilde{\bm{x}}^{n+1}$. Therefore, such an adaptive design can re-adjust the possible constraint violation by itself, which results in a regret bound of a lower order at $O(1/(N-n+1))$.

\section{Instance-dependent Sample Complexity}
In this section, we analyze the sample complexity of our \Cref{alg:Twophase}. \blue{On the correct-basis event $\mathcal B_{\mathrm I}$,
$\mathcal I^N=\mathcal I^*$ and
$\mathcal J^N=\mathcal J^*$. We use the population notation
$\mathcal I^*,\mathcal J^*$ throughout the conditional
resolving-phase analysis. Define the conditional mean output $\bm x_{\mathrm{good}}^N:=\mathbb E\left[\bar{\bm x}^N\,\middle|\,\mathcal B_{\mathrm I}\right].$} We aim to solve the LP \eqref{lp:matrix}, with the constraint matrix $A$ unknown and need to be estimated from the data. Our \Cref{alg:Twophase} is developed to solve the LP near-optimally in a data driven way. Note that after \Cref{alg:Idenbasis}, we identify the optimal basis $\mathcal{I}^*$ and $\mathcal{J}^*$. Our next lemma shows that in order to bound the gap of solving the LP \eqref{lp:matrix}, it is sufficient to analyze the term $\bc^{\blue{N+1}}_{\mathcal{J}^*}$, which is defined in \eqref{eqn:UpdateAlpha2}.

\blue{Appendix~\ref{pf:Lemma3} relates the objective residual to the terminal remaining-capacity vector. We therefore analyze the normalized process $\widetilde{\bm c}_{\mathcal J^*}(n)=\frac{\bm c_{\mathcal J^*}^n}{N-n+1}.$} Therefore, it suffices to analyze how $\bm{c}_{\mathcal{J}^*}^n$ behave. We define
\begin{equation}\label{eqn:Average}
\tilde{c}_{(s,a)}(n)=\frac{c^n_{(s,a)}}{\blue{N-n+1}},~\forall (s,a)\in\mathcal{J}^*\blue{,~n\in[N]}.
\end{equation}
The key is to show that the stochastic process $\tilde{c}_{(s,a)}(n)$ possesses some concentration properties such that they will stay within a small neighborhood of their initial value $c(s,a)$ for a sufficiently long time. We denote by $\tau$ the time that one of $\tilde{c}_{(s,a)}(n)$ for each $(s,a)\in\mathcal{J}^*$ escape this neighborhood. Then, both the gap over the objective value and the gap over the constraint satisfaction can be upper bounded by \blue{a constant times} $\blue{1+}\mathbb{E}[\blue{N+1-\tau}]$. From the update rule \eqref{eqn:UpdateAlpha2}, \blue{for $(s,a)\in\mathcal J^*$ and $n\in[N-1]$,} we know that
\begin{equation}
\tilde{c}_{(s,a)}(n+1) =
\tilde{c}_{(s,a)}(n)-\frac{A^n_{(s,a), :}\cdot\bx^n_{\mathcal{I}^*}-\tilde{c}_{(s,a)}(n)}{\blue{N-n}}. \label{eqn:Aveupalpha}
\end{equation}
Ideally, $\tilde{c}_{(s,a)}(n+1)$ will have the same expectation as $\tilde{c}_{(s,a)}(n)$ such that it becomes a martingale, for each $(s,a)\in\mathcal{J}^*$. However, this is not true since we have estimation error over $A_{\mathcal{J}^*, \mathcal{I}^*}$, and we only use their estimates to compute $\bm{x}^n$. Nevertheless, \blue{we decompose $\tilde c_{(s,a)}(n)$ into a martingale component and a controlled drift, yielding a bound on $\mathbb E[\bc_{\mathcal J^*}^{N+1}]$ for each $(s,a)\in\mathcal{J}^*$. The constraints indexed by $k\in[K]\setminus\mathcal J^*$ can be bounded using the optimal-basis structure. For completeness, if the empirical LP does not return a finite optimal basic solution, we define the final output to be $\bar{\bm x}^N=\bm0$.}


\begin{blueblock}
In the remainder of this section and in Appendices~\ref{pf:Lemma3}--~\ref{app:J}, we condition
on the correct-basis event $\mathcal B_{\mathrm I}$ and suppress this
conditioning in the probability and expectation notation. The sample cost and success
probability of basis identification are added back in
\Cref{thm:sample}. Let $B:=A_{\mathcal J^*,\mathcal I^*}$ and we define:
\begin{equation}
\sigma:=\frac{1}{\|B^{-1}\|_1}.
\label{def:sigma}
\end{equation}
Since $B$ is nonsingular, it holds that $\sigma>0$. For compactness, define
\begin{equation}\label{eqn:RN}
R_N
:=
C_c\left(N_0'+d_2G+N\epsilon_{\mathrm{II}}+1\right),
\end{equation}
where $C_c$, $N_0'$, $G$, and $\epsilon_{\mathrm{II}}$ are specified in Appendix~\ref{pf:Thm2}. We also define
\[
\Lambda
:=
\max\left\{
\|\bm y_{\mathcal J^*}^*\|_1,\,
1,\,
\max_{k\in[K]\setminus\mathcal J^*}
\left\|
A_{k,\mathcal I^*}
\left(A_{\mathcal J^*,\mathcal I^*}\right)^{-1}
\right\|_1
\right\},
\]
with the convention that the final maximum is zero when
$\mathcal J^*=[K]$. For compactness, let $\epsilon_N^{\mathrm{res}} := \frac{\Lambda R_N}{N},\bm x_{\mathrm{good}}^N:=
\mathbb E
\left[
\bar{\bm x}^N
\,\middle|\,
\mathcal B_{\mathrm I}
\right].$

\begin{theorem}\label{thm:BoundRe}
Under the standing assumptions, there exists a finite instance-dependent threshold $N_{\mathrm{res}}$ such that, for every $N\geq N_{\mathrm{res}}$,
\[
V^{\mathrm{LP}}
-
\bm r^\top\bm x_{\mathrm{good}}^N
\leq
\epsilon_N^{\mathrm{res}},
\qquad
A\bm x_{\mathrm{good}}^N-\bm c
\leq
\epsilon_N^{\mathrm{res}}\bm1.
\]
\end{theorem}
\end{blueblock}

\begin{blueblock}
Since $\epsilon_{\mathrm{II}}=1/(2N)$, we have $L_{\mathrm{II}}=2\log(2Nd_2)$ and $N_0' \leq \frac{16d_2^2}{\sigma^2}\log(2Nd_2)+1.$
Define the finite instance-dependent constant
\[
C_{\mathrm{res}}
:=
\Lambda C_c
\left(
\frac{16d_2^2}{\sigma^2}
+d_2G
+\frac{5}{2}
\right).
\]
Then, for every $N\geq N_{\mathrm{res}}, ~\Lambda R_N
\leq
C_{\mathrm{res}}
\bigl(1+\log(2Nd_2)\bigr),$
and hence
\[
\epsilon_N^{\mathrm{res}}
\leq
\frac{C_{\mathrm{res}}
\bigl(1+\log(2Nd_2)\bigr)}{N}
=
\widetilde O\left(\frac{C_{\mathrm{res}}}{N}\right).
\]
The definition in \eqref{def:sigma} is consistent with the matrix $1$-norm used in Appendix~\ref{pf:Thm2}, and we know that
\[
\|B^{-1}\|_1=\frac{1}{\sigma},
\qquad
\kappa_1(B)=\|B\|_1\|B^{-1}\|_1=\frac{\|B\|_1}{\sigma}.
\]
Thus, $\sigma$ measures the stability of the basis linear system under
the perturbations in both the estimated matrix and the normalized
remaining-capacity vector. We retain
$\|\bm y_{\mathcal J^*}^*\|_1$ explicitly in the objective bound rather
than attempting to control the dual solution by $\sigma$ alone.
\end{blueblock}

We present the final sample complexity bound of our algorithm. Note that we have a bound on the constraint violation for each $(s,a)\in\mathcal{J}^*$. For the constraints not in the index set $\mathcal{J}^*$, we can also bound its violation using the non-singularity of the matrix $A_{\mathcal{J}^*, \mathcal{I}^*}$. This is another benefit of identifying the optimal basis $\mathcal{I}^*$ and $\mathcal{J}^*$. 

\begin{newrevblock}
Define the failure-event envelope
\begin{equation}\label{eqn:Cfail}
C_{\mathrm{fail}}
:=
\max\left\{
|V^{\mathrm{LP}}|+\|\bm r\|_\infty C,\,
C+1
\right\}.
\end{equation}
For $\delta_{\mathrm I}\in(0,1)$, define $\epsilon_N^{\mathrm{mean}}:=\epsilon_N^{\mathrm{res}}+\delta_{\mathrm I}C_{\mathrm{fail}}.$
Under Assumption~\ref{assump:FunctionApp} and the projection in Algorithm~\ref{alg:Twophase}, $C_{\mathrm{fail}}$ bounds both the objective shortfall and every positive signed constraint residual on the incorrect-basis event.

\begin{theorem}
\label{thm:sample}
Let $\delta_{\mathrm I}\in(0,1)$ and choose
$N_{\mathrm I}$ such that $\operatorname{Rad} \left( N_{\mathrm I}, \frac{\delta_{\mathrm I}}{d_1K} \right) \leq \Delta_{\mathrm{id}}.$ For every $N\geq N_{\mathrm{res}}$, the mean output $\bm x_{\mathrm{mean}}^N = \mathbb E[\bar{\bm x}^N]$ satisfies
\[
V^{\mathrm{LP}}
-
\bm r^\top\bm x_{\mathrm{mean}}^N
\leq
\epsilon_N^{\mathrm{mean}},
\quad
A\bm x_{\mathrm{mean}}^N-\bm c
\leq
\epsilon_N^{\mathrm{mean}}\bm1.
\]
\end{theorem}
We now instantiate the parameters in \Cref{thm:sample} to obtain the total transition-query complexity. Algorithm~\ref{alg:Idenbasis} uses $KN_{\mathrm I}$ historical transition queries. Algorithm~\ref{alg:Twophase} uses one initial transition and one fresh transition in every resolving round for each selected row. Therefore, its exact transition-query count is
\[
Q_{\mathrm{total}}
=
KN_{\mathrm I}
+
\widehat d_2(N+1),
\quad
\widehat d_2
:=
|\mathcal I^N|
=
|\mathcal J^N|.
\]
On the correct-basis event $\mathcal B_{\mathrm I}$, $\widehat d_2=d_2$, while deterministically $\widehat d_2\leq d_1$. Let $C_{\mathrm{inst}}:=C_{\mathrm{res}}+C_{\mathrm{fail}}+N_{\mathrm{res}}.$
\begin{corollary}
\label{cor:query}
Set $\delta_{\mathrm I}=1/(2N)$.
For fixed instance parameters, $N_{\mathrm I} =\widetilde O(\Delta_{\mathrm{id}}^{-2})$ and $N =\widetilde O(C_{\mathrm{inst}}/\epsilon)$ suffice for the $\epsilon$-guarantee in Theorem~\ref{thm:sample}. Moreover, on $\mathcal B_{\mathrm I}$, which has probability at least
$1-1/(2N)$,
\[
Q_{\mathrm{total}}
=
\widetilde O
\left(
\frac{K}{\Delta_{\mathrm{id}}^2}
+
\frac{d_2C_{\mathrm{inst}}}{\epsilon}
\right).
\]
\end{corollary}

When the state-action space is large or infinite, we first construct
the RLP in \eqref{lp:Reduced} and then apply
\Cref{alg:Idenbasis,alg:Twophase} to this finite LP.

\begin{corollary}\label{coro:Reduced}
Under the constraint-sampling guarantee in
Theorem~\ref{thm:ConstSample}, the mean output is
$\epsilon$-optimal and $\epsilon$-feasible in the sense of Theorem~\ref{thm:sample}. Moreover, on the correct-basis event $\mathcal B_{\mathrm I}$, which has probability at least $1-\frac{1}{2N}$ under the choice in \Cref{cor:query}, the transition-query complexity is
\[
\widetilde O\left(
\frac{1}{\Delta_{\mathrm{id}}^2\epsilon}
+
\frac{d_2C_{\mathrm{inst}}}{\epsilon}
\right).
\]
A deterministic upper bound is obtained by replacing $d_2$ with $d_1$ in the second term.
\end{corollary}
\end{newrevblock}

\blue{A key quantity in our sample-complexity bound is the LP suboptimality gap $\Delta$, which differs from the $Q$-value gap commonly used in prior work (e.g., \cite{simchowitz2019non,he2021logarithmic}):}
\begin{equation}\label{def:deltaminimal}
\delta := \min_{(s,a)\in\mathcal{S}\times\mathcal{A}:\, Q^*(s,a)<V^*(s)} \left\{V^*(s)-Q^*(s,a)\right\}.
\end{equation}
\blue{This quantity measures the smallest gap of any suboptimal action and may equal zero when $\mathcal S$ is infinite, so previous analyses typically assume $\delta>0$. In contrast, our $\Delta$ measures the separation between optimal LP basic solutions and infeasible or suboptimal ones. Our $\Delta$-dependent results require only that the two minimization domains in \eqref{def:delta} are not both empty, without imposing a uniform positive $Q$-value gap over all states.}

\begin{newrevblock}
We next state a population-level policy consequence of the mean-iterate guarantee. Define $V_{\mathrm{mean}}^N(s) = \Phi(s)^\top\bm x_{\mathrm{mean}}^N$, $s\in\mathcal S,$ and the oracle greedy policy
\begin{equation}\label{eq:policy_mean}
\hspace{-0.1cm}
\pi_{\mathrm{mean}}^N(s)
\in
\arg\min_{a\in\mathcal A}
\left\{
c(s,a)
+
\gamma
\mathbb E_{s'\sim P(\cdot\mid s,a)}
\left[
V_{\mathrm{mean}}^N(s')
\right]
\right\}
\end{equation}
The true transition kernel is used only for this theoretical
implication. Constructing a policy using an estimated kernel requires
a separate policy-extraction error analysis.

\begin{corollary}
\label{cor:value_function_policy}
Suppose that \Cref{thm:sample} applies to the full ALP and, for some $\bm x^*$,
\[
V^{\pi^*}(s)=\Phi(s)^\top\bm x^*,
\quad
V^{\mathrm{LP}}
=
\int_{\mathcal S}V^{\pi^*}(s)\,\mu(ds),
\quad
\mu(\mathcal S)=1.
\]
If $\epsilon_N^{\mathrm{mean}}\leq\epsilon$, then, for every
$\nu$ with
$C_{\nu,\mu}^{\pi_{\mathrm{mean}}^N}<\infty$,
\[
V^{\pi_{\mathrm{mean}}^N}(\nu)-V^{\pi^*}(\nu)
\leq
\frac{(2-\gamma)
C_{\nu,\mu}^{\pi_{\mathrm{mean}}^N}}
{(1-\gamma)^2}\epsilon.
\]
\end{corollary}

This corollary concerns the oracle policy induced by the
mean iterate. The expected LP guarantee in \Cref{thm:sample} does not,
by itself, imply the same policy guarantee for a greedy policy formed
from one realized output $\bar{\bm x}^N$.
\end{newrevblock}

\section{Concluding Remarks}

\begin{blueblock}
In this work, we propose a new LP-based resolving algorithm for MDPs with function approximation. Conditional on successful basis identification, we prove that the expected averaged iterate after $N$ resolving rounds achieves an instance-dependent $\widetilde O(C_{\mathrm{inst}}/N)$ objective shortfall and signed constraint residual. We also distinguish resolving rounds from individual transition queries: Algorithm~\ref{alg:Twophase} uses $d_2$ fresh transitions per round, while the historical samples used by Algorithm~\ref{alg:Idenbasis} are counted separately. Combining the two phases gives the total transition-query bound in \Cref{cor:query}. We formulate the problem as an LP, use function approximation for tractability, and exploit the identified basis to reduce the dimension of the learned linear system. The experiments illustrate the practical benefit of adaptive resolving. We focus on the generative-model setting and leave extensions to offline and online RL for future work.
\end{blueblock}
\bibliographystyle{abbrvnat}
\bibliography{bibliography}

\clearpage

\OneAndAHalfSpacedXI

%
%
%

\begin{APPENDICES}
\crefalias{section}{appendix}

\section{Numerical Experiments}\label{sec:numerical}
We conduct numerical experiments to test the practical performance of our algorithms. We test on the ``Mountain Car'' problem, which is a well-known continuous control task, with additional random noise to represent the sampling uncertainty in real-life scenarios. The detailed experimental setup is reported in \Cref{sec:ExperimentSetup} below and the numerical results are presented in \Cref{sec:NumericalResults}.

\subsection{Experimental Setup}\label{sec:ExperimentSetup}
We report the detailed experimental setup of our numerical experiments.

\textbf{Mountain Car Problem}.
The ``Mountain Car'' problem is a well-known continuous control task. The car is located in the middle of the valley and we need to control the acceleration of the car so that it can overcome its gravity and climb to the top of the mountain.
It has a state space $S \subset R^2$ and an action space $A \subset R$, where the state $s$ is made up of position $p$ and velocity $v$. The domain of these variables are:
$$p \in [-1.20, 0.60], \hspace{5pt} v \in [-0.07, 0.07], \hspace{5pt} a \in [-1, 1].$$
The car intends to follow the following transition dynamics:
\begin{equation}\label{eqn:dynamics}
    \begin{aligned}
        v_{t+1} &= v_t + a * 0.0015 - 0.0025 * cos(3 * p_t), \\
        p_{t+1} &= p_t + v_{t+1}.
    \end{aligned}
\end{equation}
However, instead of letting the car follow the transition dynamics in \eqref{eqn:dynamics} deterministically, there are some additional random noises in our simulator, which represents the sampling uncertainty in real-life scenarios and makes our setting more challenging. To be specific, we add a random noise $\epsilon$ to the simulator to simulate various unexpected situations in reality. Define the function $K: S \times A \rightarrow S$ to represent the intended state transition given state $s=(p,v)$ and action $a$ as described in \eqref{eqn:dynamics}. Then, the probability of the next state $s^{'}$ given current state $s$ and action $a$ is:
\begin{equation}
    P(s^{'}|s,a) = \left\{
    \begin{aligned}
    & 0.9 + 0.1 * P_{random}(s^{'}), \hspace{1pt} \text{if} \hspace{1pt} s^{'} == K(s, a)\\
    & 0.1 * P_{random}(s^{'}), \hspace{1pt} \text{otherwise}
    \end{aligned}
    \right.
\end{equation}
$P_{random}(s^{'})$ is the probability that $s^{'}$ is uniformly distributed in the neighborhood of the real next state $K(s,a)$. The neighborhood can be described as an interval. Denote by $i^{\mathrm{intend}}$ the index of $K(s,a)$, then the neighborhood is given by $[i^{\mathrm{intend}} - r_{\epsilon}, i^{\mathrm{intend}} + r_{\epsilon}]$, where $r_{\epsilon}$ is a pre-specified range. The real transition procedure of our setting can be described as with probability $0.9$, the next state is $K(s,a)$, and with probability $0.1$, the next state is uniformly sampled from a small neighborhood of $K(s,a)$ with a radius $r_{\epsilon}$.


\textbf{Linear Function Approximation}.
We approximate the value function by defining feature maps for position
$p$ and velocity $v$ separately. Specifically, we use
\blue{radial basis functions}. Let $p_i$ and $v_i$, $i\in[5]$, be
equally spaced centers in the position and velocity domains, and define
\begin{blueblock}
\begin{equation}
\begin{aligned}
&\phi_p(p)
=
\left(
\exp\!\left[-\left(\frac{p-p_i}{0.2}\right)^2\right]
\right)_{i=1}^5
\in\mathbb R^5,\\
&\phi_v(v)
=
\left(
\exp\!\left[-\left(\frac{v-v_i}{0.2}\right)^2\right]
\right)_{i=1}^5
\in\mathbb R^5.
\end{aligned}
\end{equation}
The overall feature vector is the vectorized outer product
\begin{equation}
\phi_V(p,v)
:=
\operatorname{vec}\!\left(
\phi_p(p)\phi_v(v)^\top
\right)
\in\mathbb R^{25}.
\end{equation}
\end{blueblock}
Given any position $p$, velocity $v$ and weight $w$, we can get the V-function: $f_w(p, v) := \langle w,\phi_V(p,v)\rangle$.

\textbf{Experiment Configuration}.
Note that ``Mountain Car'' problem is a continuous control problem with infinitely large state-action space, thus the ALP \eqref{lp:matrix} is a semi-infinite LP with infinite number of constraints. In order to solve the ALP \eqref{lp:matrix} approximately, we adopt the idea of constraint sampling described in \Cref{sec:extension} to consider the reduced LP (RLP) as given in \eqref{lp:Reduced}. The constraint sampling procedure can be regarded as discretizing the state-action space. To be specific, our constraint sampling procedure can be regarded as dividing position $p$, velocity $v$ and action $a$ into 40, 60 and 5 parts, respectively. Therefore, there are in total $12,000$ constraints in the reduced LP (RLP) in \eqref{lp:Reduced}. For our resolving algorithm, we apply \Cref{alg:Idenbasis} and \Cref{alg:Twophase} to the RLP \eqref{lp:Reduced} to approximate the optimal weight. For the non-resolving algorithm, we use the same number of samples to construct an estimation of the RLP \eqref{lp:Reduced} and directly solve this estimated RLP to approximate the optimal weight. 

We define that $y_{real}$ refers to the optimal value of the benchmark RLP \eqref{lp:Reduced} and $y_{resolve}$ is the objective value of the RLP under our solution, then the relative optimal value gap is:
$$\frac{|y_{real} - y_{resolve}|}{y_{real}}.
$$

We define that $\bx_{real}$ refers to the solution of the benchmark RLP \eqref{lp:Reduced} corresponding to the optimal basis identified by \Cref{alg:Idenbasis}, and $\bx_{resolve}$ is our solution, then the relative solution gap is:
$$\frac{\|\bx_{real} - \bx_{resolve}\|_2}{\|\bx_{real}\|_2}.
$$

After substituting our result $\bx_{resolve}$ into constraints, constraint violation refers to the maximum constraint as they are expected to be equal or less than zero:
\begin{equation}
\max\left\{([A \cdot \bx_{resolve} - \bc]^+)\right\}.\nonumber
\end{equation}

When constructing the RLP \eqref{lp:Reduced} and the dataset $H^n$ in Algorithm \ref{alg:Twophase}, for each $(s,a)$ pair, we sample $L$ number of state transition $s^{'}$ from the simulator as its next state. Therefore, suppose $n_{s^{'}}$ represents the sampling number of state $s^{'}$, the probability of each state is:
$$P(s^{'}|(s,a))=\frac{n_{s^{'}
}}{L}
$$


The true RLP is the benchmark and the output of our algorithm should be close to the true RLP. We evaluate this from three aspects: relative optimal value gap, constraints violations and the relative solution gap, with the definitions given above. In the implementation of our algorithm (\Cref{alg:Idenbasis} and \Cref{alg:Twophase}), we use the first $120,000$ number of samples ($10$ samples for each constraint) to construct an estimation of the RLP and then carry out \Cref{alg:Idenbasis} to identify the basis $\mathcal{I}^*$ and $\mathcal{J}^*$. We then use the rest of the samples to carry out the resolving steps in \Cref{alg:Twophase}. In the implementation of the non-resolving algorithm, we use the same amount of samples to construct an estimate of the RLP, and directly solve this estimation to obtain a solution. Note that if a total $240,000$ samples are used in our algorithm ($120,000$ for \Cref{alg:Idenbasis} and $120,000$ for \Cref{alg:Twophase}), we also use the same amount $240,000$ samples for the non-resolving algorithm, with $20$ sample for each constraint as the original RLP has $12,000$ constraints.

\subsection{Experiment Results}\label{sec:NumericalResults}
\begin{figure}[htbp]
    \centering
    \includegraphics[width=0.5\textwidth]{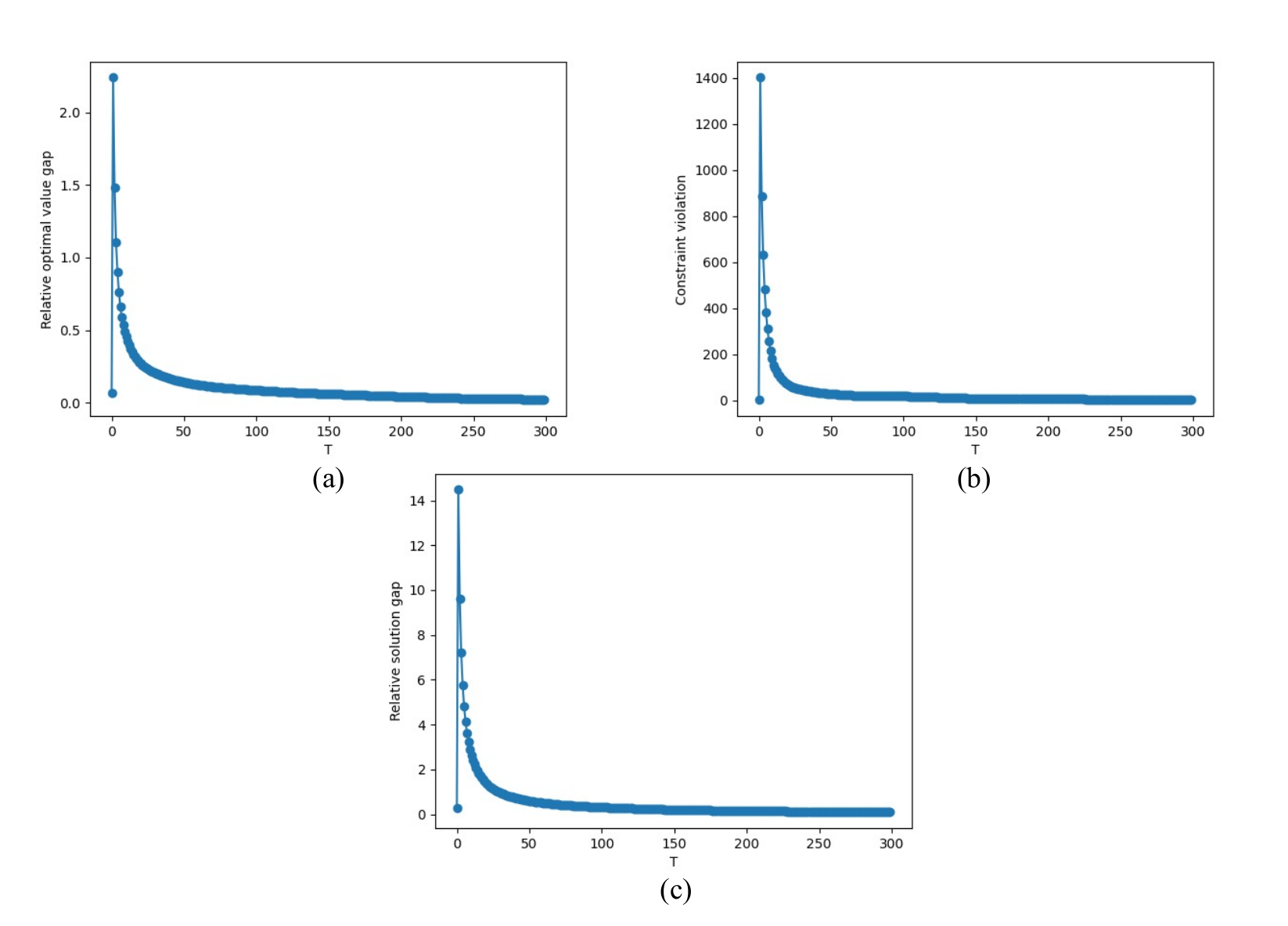}
    \caption{Numerical performance of our resolving algorithm on the Mountain Car Problem. (a) The relative optimal value gap between our algorithm \ref{alg:Twophase} and the benchmark RLP \eqref{lp:Reduced}. (b) The maximum constraint violation after substituting our result into constraints. If the constraints are satisfied, then the violation is $0$. (c) The relative gap between our LP solution and the real LP solution.  }
    \label{fig:mountain_car_results}
\end{figure}

\begin{figure}[htbp]
    \centering
    \includegraphics[width=0.5\textwidth]{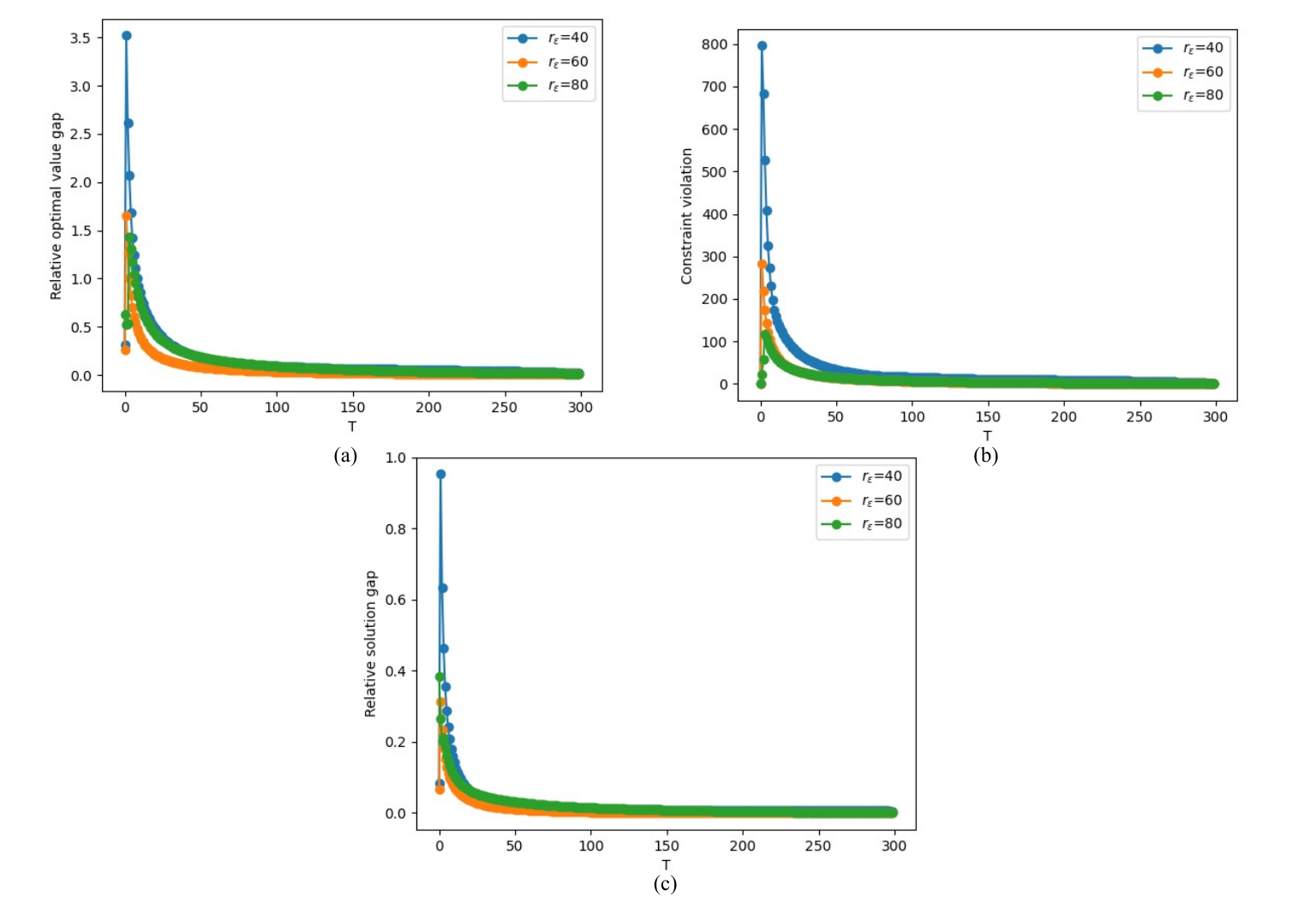}
    \caption{Numerical performance comparison between different random noise radius $r_{\epsilon}$. (a) The relative optimal value gap between our algorithm \ref{alg:Twophase} and the benchmark RLP \eqref{lp:Reduced}. (b) The maximum constraint violation after substituting our result into constraints. If the constraints are satisfied, then the violation is $0$. (c) The relative gap between our LP solution and the real LP solution.}
    \label{fig:results_comparison}
\end{figure}

\begin{figure}[htbp]
    \centering
    \includegraphics[width=0.48\textwidth]{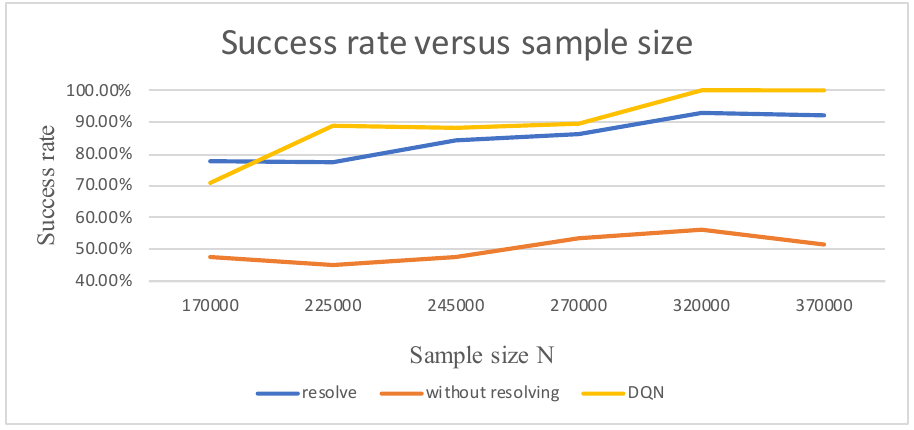}
    \caption{Success rates of our algorithm on the real Mountain Car Problem. We compare our resolving algorithm with a Deep Q-learning Network and a non-resolving baseline that directly solves the estimated LP, under the same total transition-query budget.}
    \label{fig:success_rate}
\end{figure}

We use $T$ to denote the number of resolving steps in our algorithm. We study how our algorithm converge with the number of resolving steps.
At $T = 0$, our resolving algorithm has not started yet and the estimated LP is still far from the real LP. 
\Cref{fig:mountain_car_results} (a) and (c) show that the solution without resolving is still far from the benchmark, which reveals the necessity and importance of our resolving algorithm. As $T$ increases with more and more resolving steps, as shown in \Cref{fig:mountain_car_results} (a), the relative optimal value gap converges to $0$ as $T$ increases, which means our resolving algorithm's performance is close to the benchmark. In \Cref{fig:mountain_car_results} (c), 
the relative error of our LP solution and the true LP solution also converges to zero. These two figures together confirm that our resolving algorithm obtains a result that is very close to the benchmark. \Cref{fig:mountain_car_results} (b) further tests whether our solution satisfies the constraints, and the maximum constraint violation error converges to 0, indicating that our solution can approximately satisfy all constraints. The exact definitions of the metrics in \Cref{fig:mountain_car_results} are presented in \Cref{sec:ExperimentSetup}.

\Cref{fig:results_comparison} studies the performances of our algorithm with respect to different noise radius $r_{\epsilon}$ and all three figures share common trends. The setting with $r_{\epsilon}=40$ has the smallest range of the random noise, while $r_{\epsilon}=80$ has the largest range. When $r_{\epsilon}$ decreases, convergence rates in all three figures become faster. This is easy to interpret as that a smaller $r_{\epsilon}$ indicates a smaller range of random noise, which makes it easier for our algorithm to converge to the real LP solution. 

Moreover, we implement our policy on the real Mountain Car problem to see whether our policy can succeed in real problems. Following classical settings, we restrict our policy to push the car to reach the top of the mountain within $1000$ steps. We repeat the experiments $1000$ times and we compute the success rate, i.e., the percentage of times that our policy can push the car to reach the top of the mountain within $1000$ steps. We compare the performances of our resolving algorithm (\Cref{alg:Idenbasis} and \Cref{alg:Twophase}) with the Deep Q-learning Network that has the same number of parameters as our resolving algorithm and a non-resolving algorithm that directly uses the same amount of sample to construct the estimation of the ALP \eqref{lp:matrix} or the RLP \eqref{lp:Reduced} to approximate the LP solution.
\Cref{fig:success_rate} shows a significant \blue{improvement} in the success rate of our algorithm compared to the non-resolving algorithm.
\Cref{fig:success_rate} shows that with the same amount of sample $N$, the performance of our algorithm is significantly improved. The success rate of the non-resolving algorithm is about $50\%$, and the success rate is increased by about $40\%$ with our algorithm.
In addition, as the sample size increases, the success rate of our algorithm can reach up to $92.5\%$, which reveals the great empirical performances of our algorithm for solving real-life problems.
Besides, \Cref{fig:success_rate} also shows that the performance of our algorithm is comparable to DQN, where the difference between the success rate is relatively small, and our algorithm even outperforms DQN when the number of samples is relatively small (about 170,000 samples). We believe these results demonstrate the effectiveness of our algorithm.

\textbf{Discussions on computation cost}. Note that our \Cref{alg:Idenbasis} can be conducted very efficiently in practice, and the computational complexity is equivalent to solving the LP by one time, using the simplex method. In our numerical experiment, we use Gurobi (as well as COPT) to solve the reduced LP, which has 12,000 constraints, and it takes $0.3$ second to finish running \Cref{alg:Idenbasis}.  We run our code on a computer with Apple M1 chip, 8 GB memory, and it only takes about one hour to run the entire \Cref{alg:Idenbasis} and \Cref{alg:Twophase}, which highlights the numerical efficiency of our approach. Moreover, we can also adopt \Cref{alg:Idenbasis_original} in \Cref{sec:Basis} to identify the basis. Note that \Cref{alg:Idenbasis_original} only needs to solve an LP by a finite number of times and also can be carried out very efficiently due to the power of modern LP solvers. Moreover, it has been demonstrated that \Cref{alg:Idenbasis_original} enjoys a polynomial-time computational complexity. In this way, we conclude that our algorithms are numerically efficient.


\section{Discussions on the Generative Model Assumption}
Generative-model access is a common assumption in the MDP and RL
literature \citep{sidford2018near,yang2019sample,liu2025sample}, while
other works collect data through trajectories
\citep{song2022hybrid,jin2023provably}.
\begin{blueblock}
Trajectory data can support the same matrix-estimation procedure when
the data-collection policy provides adequate coverage of the
state-action pairs used by the finite LP. Without such a coverage
condition, trajectory access is generally weaker than generative-model
access. We use a simulator in the experiments to match the
generative-model assumption in our analysis and to isolate the
statistical behavior of the resolving procedure.
\end{blueblock}

\section{Additional Algorithm for Basis Identification}\label{sec:Basis}

Here we introduce an additional method to identify the basis of LP variables and constraints, which is also based on the \Cref{lem:Basis}. This algorithm solves the LPs by $K+d$ times and thus enjoys a polynomial-time worst-case computational complexity.

Note that if a variable is not a basic variable, we can restrict its value to $0$ without changing the LP values.
To detect whether we can restrict one variable to be $0$ without changing the LP value, we can add the constraint $x_i=0$ to LP \eqref{lp:matrix} and compare its value to the original LP. If the objective value has not changed, then we know that the $i$-th variable is not a basic variable and does not belong to the basis $\mathcal{I}^*$. We repeat the above procedure for each variable $i$. Note that during the repeating procedure, if we can restrict one variable $x_i=0$ without changing the LP value, we will remain this restriction when we test the remaining variables. In this way, we identify one optimal basis from possibly many. To this end, for an index set $\mathcal{I}$, we define the following LP with the variables not in $\mathcal{I}$ restricted to be $0$, as well as its estimate,
\begin{blueblock}
\begin{equation}\label{lp:Izero}
\begin{alignedat}{4}
V^{\mathrm{LP}}_{\mathcal I}
&=\max\limits_{\bm x\in\mathbb R^{d_1}} &~& \bm r^\top\bm x
&
\widehat V^{\mathrm{LP}}_{\mathcal I}(\mathcal H)
&=\max\limits_{\bm x\in\mathbb R^{d_1}} &~& \bm r^\top\bm x\\
&\mathrm{s.t.}&& A\bm x\leq\bm c
&
&\mathrm{s.t.}&& \widehat A(\mathcal H)\bm x\leq\bm c\\
&&& \bm x_{\mathcal I^c}=0
&
&&& \bm x_{\mathcal I^c}=0.
\end{alignedat}
\end{equation}
\end{blueblock}

\begin{algorithm}[ht!]
\caption{Additional algorithm for optimal basis identification}
\label{alg:Idenbasis_original}
\begin{algorithmic}[1]
\State \textbf{Input:} the historical sample set $\mathcal{H}$ with $N$ transition data for each $(s,a)$.
\State Compute the value of $\hat{V}^{\mathrm{LP}}(\mathcal{H})$ as in \eqref{lp:Izero}.

\State Initialize $\mathcal{I}=[d_1]$ to be the whole index set that contains every variable of LP \eqref{lp:matrix} and $\mathcal{J}=\mathcal{K}$.

\For{$i\in\mathcal{I}$}
\State Let $\mathcal{I}'=\mathcal{I}\backslash\{i\}$.
\State Compute the value of $\hat{V}^{\mathrm{LP}}_{\mathcal{I}'}(\mathcal{H})$ as in \eqref{lp:Izero}.
\State If $|\hat{V}^{\mathrm{LP}}(\mathcal{H})-\hat{V}^{\mathrm{LP}}_{\mathcal{I}'}(\mathcal{H})|\leq\sqrt{\Rad(N, \blue{\eps_0})}$ with $\Rad(N, \blue{\eps_0})$ given in \eqref{eqn:Rad}, then we set $\mathcal{I}=\mathcal{I}'$.
\EndFor
\For{$(s,a)\in\mathcal{J}$}
\State Let $\mathcal{J}'=\mathcal{J} \backslash\{(s,a)\}$.
\State Compute the value of $\hat{D}^{\mathrm{LP}}_{\mathcal{I}, \mathcal{J}'}(\mathcal{H})$ as in \eqref{lp:Jzero}.
\State If $|\hat{V}^{\mathrm{LP}}(\mathcal{H})-\hat{D}^{\mathrm{LP}}_{\mathcal{I}, \mathcal{J}'}(\mathcal{H})|\leq\sqrt{\Rad(N, \blue{\eps_0})}$, then we set $\mathcal{J}=\mathcal{J}'$.
\EndFor
\State \textbf{Output}: the sets of indices $\mathcal{I}$ and $\mathcal{J}$.
\end{algorithmic}
\end{algorithm}

We also need to detect whether a constraint is binding under the optimal basic solution corresponding to the basis $\mathcal{I}^*$. In order to tell whether a constraint is redundant and can be removed (not in the index set $\mathcal{J}^*$), we can consider the dual of LP \eqref{lp:matrix}, with the additional constraints $\bm{x}_{\mathcal{I}^{*c}}=0$. If a dual variable can be restricted to $0$ without influencing the LP value, we know that the corresponding constraint is redundant. Denote by $\mathcal{J}$ the index set of the dual variables that could be restricted to $0$. 
The dual LP with the restriction set $\mathcal{J}$ can be written as follows, as well as its empirical estimation.
Denote by $\mathcal{J}$ the index set of the constraints that could be binding. We consider the following LP with the constraints not in $\mathcal{J}$ removed, as well as its estimate,

\begin{equation}\label{lp:Jzero}
\begin{aligned}
D^{\mathrm{LP}}_{\mathcal{I}, \mathcal{J}}=& \min~~\bm{c}^\top \bm{y} && \hat{D}^{\mathrm{LP}}_{\mathcal{I}, \mathcal{J}}(\mathcal{H})=\min~~\bm{c}^\top \bm{y}\\
&~~ \mbox{s.t.}~~A^\top_{:, \mathcal{I}}\bm{y}=\bm{r}_{\mathcal{I}}  && ~~~~~~~~~~~~~~~~~~~~~~~\mbox{s.t.} ~~\hat{A}^\top_{:, \mathcal{I}}(\mathcal{H})\bm{y}=\bm{r}_{\mathcal{I}}\\
& ~~~~~~~~~~\bm{y}_{\mathcal{J}^c}=0  &&~~~~~~~~~~~~~~~~~~~~~~~~~~~~~~~\bm{y}_{\mathcal{J}^c}=0  \\
& ~~~~~~~~~~\bm{y}\geq0, && ~~~~~~~~~~~~~~~~~~~~~~~~~~~~~~~\bm{y}\geq0.  \\
\end{aligned}
\end{equation}
As we can see, \Cref{alg:Idenbasis_original} only requires us to compute the LP values by a finite number of times. More specifically, we need to compute the LP values for $K+d_1$ number of times. The size of the LP is also polynomial in the dimension of the problem (the size is $K\times d_1$). Therefore, we know that
\Cref{alg:Idenbasis_original} can be conducted in polynomial time.


\section{Proof of \Cref{lem:Basis}} \label{app:D}

Our proof follows from standard LP theory regarding the optimality of
basic solutions, with mild modifications.
\begin{blueblock}
Because the weight variables are unrestricted, the basis can be
interpreted using a generalized simplex implementation or, equivalently,
after representing each unrestricted variable as the difference of two
nonnegative variables. The index set $\mathcal I$ below refers to the
corresponding original weight variables.
\end{blueblock}
Note that in the LP standard form
\begin{equation} \label{lp:newmatrix}
V^{\mathrm{LP}}=~\max~ \bm{r}^\top \bm{x} ~~~ \mathrm{s.t.} ~A\bm{x} +\bm{s} = \bm{c},~~~ \bm{x}\in\mathbb{R}^{d_1}, \bm{s}\geq0,
\end{equation}
we can sort the decision variable as $(\bx, \bm{s})\in\mathbb{R}^{d_1+K}$ and we let an index set $\mathcal{I}'\subset[d_1+K]$ be an optimal basis set (there must exist an optimal basis set). Then, from the standard LP theory, the following condition holds for the optimal basis set $\mathcal{I}'$. It holds $|\mathcal{I}'| = K$. Also, we can divide $\mathcal{I}'$ as $\mathcal{I}'=\mathcal{I}'_1\cup\mathcal{I}'_2$ with $\mathcal{I}'_1$ being an index set for the $\bx$ variable and $\mathcal{I}'_2$ being an index set for the $\bm{s}$ variable. Then, we have that the solution $(\bx^*, \bm{s^*})$ satisfying the conditions
\begin{equation}\label{eqn:New001}
\bx^*_{\mathcal{I}^{'c}_1} = 0, ~~~~\bm{s}^*_{\mathcal{I}^{'c}_2} = 0
\end{equation}
and
\begin{equation}\label{eqn:New002}
A_{:, \mathcal{I}_1}\cdot \bx^*_{\mathcal{I}_1} + \bm{s}^* = \bc.
\end{equation}
is an optimal solution to $V^{\mathrm{LP}}$. 
Moreover, the linear system described in \eqref{eqn:New001} and \eqref{eqn:New002} is uniquely determined. We now write the linear system in \eqref{eqn:New001} and \eqref{eqn:New002} into the matrix form. Since $|\mathcal{I}'| = K = |\mathcal{I}'_1| + |\mathcal{I}'_2|$, we define the matrix
\[
\bar{A} = \begin{bmatrix}
    A_{\mathcal{I}^{'c}_2, ~\mathcal{I}_1}, & 0, \dots, 0\\
    A_{\mathcal{I}'_2,~\mathcal{I}_1}, & I_{|\mathcal{I}'_2|}
\end{bmatrix}
\]
where $I_{|\mathcal{I}'_2|}$ denotes an identity matrix with size $|\mathcal{I}'_2|$. Then, the linear system described in \eqref{eqn:New001} and \eqref{eqn:New002} can be written as
\[
\bx^*_{\mathcal{I}^{'c}_1} = 0, ~~~~\bm{s}^*_{\mathcal{I}^{'c}_2} = 0
\]
and
\begin{equation}\label{eqn:New003}
    \bar{A}\cdot \begin{pmatrix}
        \bx^*_{\mathcal{I}_1}\\
        \bm{s}^*_{\mathcal{I}_2}
    \end{pmatrix} = \bc.
\end{equation}
Since the linear system described in \eqref{eqn:New001} and \eqref{eqn:New002} is uniquely determined, we know that the square matrix $\bar{A}$ is non-singular. 

The non-singularity of the matrix $\bar{A}$ implies the non-singularity of the matrix $A_{\mathcal{I}^{'c}_2, ~\mathcal{I}_1}$. To see this, we first note that since $\bm{s}\in\mathbb{R}^K$, we have $|\mathcal{I}'_2| + |\mathcal{I}^{'c}_2| = K$ and we also have $ |\mathcal{I}'_1| + |\mathcal{I}'_2| = K$, which implies that $|\mathcal{I}'_1| = |\mathcal{I}^{'c}_2|$ and thus the matrix $A_{\mathcal{I}^{'c}_2, ~\mathcal{I}_1}$ is a square matrix. Moreover, the non-singularity of $\bar{A}$ must imply that the rows of the matrix $A_{\mathcal{I}^{'c}_2, ~\mathcal{I}_1}$ must be linearly independent from each other. Thus, we know that the square matrix $A_{\mathcal{I}^{'c}_2, ~\mathcal{I}_1}$ is non-singular. Therefore, we know that $\bx^*_{\mathcal{I}_1}$ can be uniquely determined as the unique solution to the linear system
\begin{equation}\label{eqn:New004}
    A_{\mathcal{I}^{'c}_2, ~\mathcal{I}_1}\cdot \bx^*_{\mathcal{I}_1} = \newrev{\bc_{\mathcal{I}^{'c}_2}}.
\end{equation}
We now set the index set
\[
\mathcal{I} = \mathcal{I}'_1, \text{~~and~~}\mathcal{J} = \mathcal{I}^{'c}_2.
\]
Then, we know that 
\[
|\mathcal{I}| = |\mathcal{J}|
\]
and a solution $\bx^*$ uniquely defined as 
\[
\bx^*_{\mathcal{I}^c} = 0
\]
and 
\[
A_{\mathcal{I}^{'c}_2, ~\mathcal{I}_1}\cdot \bx^*_{\mathcal{I}_1} = \bc
\]
is an optimal solution to $V^{\mathrm{LP}}$, where $\bm{s}$ is the slackness variables determined corresponding to $\bx^*$. Our proof is thus completed.

\section{Proof of \Cref{thm:Infibasis2}} \label{app:E}
\blue{For brevity, let $\rho_N:=\operatorname{Rad}(N,\epsilon_0).$} We now condition on the event that 
\begin{equation}\label{def:event}
\begin{aligned}
\mathcal{E}=\Biggl\{\left|\widehat A_{k,i}(\mathcal H)-A_{k,i}\right|
\leq \Rad(N, \blue{\eps_0}), ~~\forall k\in[K], ~\forall i\in[d_1]  \Biggr\}.
\end{aligned}
\end{equation}
We know that this event $\mathcal{E}$ happens with probability at least \blue{$1-Kd_1\eps_0=1-\eps$}.

\blue{Recall that $\mathcal F_1$ is the finite collection of
population basis pairs
$(\mathcal I,\mathcal J)$ for which
$A_{\mathcal J,\mathcal I}$ is nonsingular. Define
\[
M_{\mathrm{bas}}
:=
\max_{(\mathcal I,\mathcal J)\in\mathcal F_1}
\max\left\{
\left\|
A_{\mathcal J,\mathcal I}^{-1}
\right\|_1,
\left\|
A_{\mathcal J,\mathcal I}^{-\top}
\right\|_1
\right\}.
\]
Since $\mathcal F_1$ is finite and contains the population
optimal basis,
\[
M_{\mathrm{bas}}<\infty.
\]
We further define
\[
\eta_{\mathrm{id}}
:=
\min\left\{
1,\,
\eta_{\mathrm{reg}},\,
\frac{1}{2d_1M_{\mathrm{bas}}}
\right\}.
\]}

We first bound the gap between the population and empirical
optimal values. The result is formalized in the following
claim, whose proof is given at the end of this section.
\begin{newrevblock}
\begin{claim}
\label{claim:Bound1}
Conditional on the event $\mathcal E$ in~\eqref{def:event},
suppose that
\[
\rho_N
\leq
\eta_{\mathrm{id}}.
\]
Then
\[
\left|
V^{\mathrm{LP}}
-
\widehat V^{\mathrm{LP}}(\mathcal H)
\right|
\leq
C_{\mathrm{val}}\rho_N,
\]
where
\[
C_{\mathrm{val}}
:=
2M_{\mathrm{bas}}B_x\|\bm r\|_1.
\]
\end{claim}
\end{newrevblock}
\begin{newrevblock}
Let $(\widehat{\mathcal I},\widehat{\mathcal J})$ be the
optimal basis returned by Algorithm~\ref{alg:Idenbasis}, and let $\widehat{\bm x}$ be its optimal basic solution. Thus,
\begin{equation}\label{eqn:New02}
\widehat{\bm x}_{\widehat{\mathcal I}^c}=0,
\qquad
\widehat A_{\widehat{\mathcal J},\widehat{\mathcal I}}(\mathcal H)
\widehat{\bm x}_{\widehat{\mathcal I}}
=
\bm c_{\widehat{\mathcal J}}.
\end{equation}
For the remainder of the proof, suppose that
\[
\rho_N
\leq
\eta_{\mathrm{reg}}.
\]
By Assumption~\ref{assump:degeneracy}, the optimal basic
solution selected by Algorithm~\ref{alg:Idenbasis} satisfies
\[
\|\widehat{\bm x}\|_1
\leq
B_x,
\]
and the population matrix
\[
A_{\widehat{\mathcal J},
  \widehat{\mathcal I}}
\]
is nonsingular.

Define the population basic solution $\bm x'$ associated with
the selected empirical basis by
\[
\bm x'_{\widehat{\mathcal I}^c}
=
\bm0,
\qquad
A_{\widehat{\mathcal J},
  \widehat{\mathcal I}}
\bm x'_{\widehat{\mathcal I}}
=
\bm c_{\widehat{\mathcal J}}.
\]
The empirical and population basis systems satisfy
\[
\widehat A_{\widehat{\mathcal J},
             \widehat{\mathcal I}}(\mathcal H)
\widehat{\bm x}_{\widehat{\mathcal I}}
=
\bm c_{\widehat{\mathcal J}}
\]
and
\[
A_{\widehat{\mathcal J},
  \widehat{\mathcal I}}
\bm x'_{\widehat{\mathcal I}}
=
\bm c_{\widehat{\mathcal J}}.
\]
Therefore,
\[
A_{\widehat{\mathcal J},
  \widehat{\mathcal I}}
\left(
\widehat{\bm x}_{\widehat{\mathcal I}}
-
\bm x'_{\widehat{\mathcal I}}
\right)
=
\left(
A_{\widehat{\mathcal J},
  \widehat{\mathcal I}}
-
\widehat A_{\widehat{\mathcal J},
             \widehat{\mathcal I}}(\mathcal H)
\right)
\widehat{\bm x}_{\widehat{\mathcal I}}.
\]
Consequently,
\[
\begin{aligned}
\|\widehat{\bm x}-\bm x'\|_1
&\leq
\left\|
A_{\widehat{\mathcal J},
  \widehat{\mathcal I}}^{-1}
\right\|_1
\left\|
A_{\widehat{\mathcal J},
  \widehat{\mathcal I}}
-
\widehat A_{\widehat{\mathcal J},
             \widehat{\mathcal I}}(\mathcal H)
\right\|_1
\|\widehat{\bm x}\|_1 \\
&\leq
M_{\mathrm{bas}}\,
|\widehat{\mathcal J}|\,
\rho_N B_x \\
&\leq
d_1M_{\mathrm{bas}}B_x\rho_N.
\end{aligned}
\]
Define
\[
C_{\mathrm{bas}}
:=
d_1M_{\mathrm{bas}}B_x.
\]
Then
\[
\|\widehat{\bm x}-\bm x'\|_1
\leq
C_{\mathrm{bas}}\rho_N.
\label{eq:local_basis_comparison}
\]
It follows that
\[
\|\bm x'\|_1
\leq
B_x+C_{\mathrm{bas}}\eta_{\mathrm{id}}.
\]
\end{newrevblock}
We now show that when $N$ is sufficiently large, $\bx'$ becomes an optimal solution to $V^{\mathrm{LP}}$. If not, we classify into the two possible situations:

\noindent Situation (i): the basic solution $\bx'$ is an infeasible solution to $V^{\mathrm{LP}}$.
From the feasibility of $\hat{\bx}$ to the empirical LP, we know that
\[
\widehat A(\mathcal H)\widehat{\bm x}
\leq
\bm c.
\]
\begin{newrevblock}
Moreover, Assumption~\ref{assump:FunctionApp} implies
\[
\|\widehat A(\mathcal H)\|_{\max}\leq1.
\]
Therefore,
\[
\begin{aligned}
A\bm x'
&=
\widehat A(\mathcal H)\bm x'
+
\left(
A-\widehat A(\mathcal H)
\right)\bm x'\\
&\leq
\widehat A(\mathcal H)\widehat{\bm x}
+
\|\widehat{\bm x}-\bm x'\|_1\bm1
+
\rho_N\|\bm x'\|_1\bm1\\
&\leq
\bm c
+
C_{\mathrm{feas}}\rho_N\bm1,
\end{aligned}
\]
where
\[
C_{\mathrm{feas}}
:=
B_x+
\left(
1+\eta_{\mathrm{id}}
\right)C_{\mathrm{bas}}.
\]
If $\bm x'$ is infeasible and the minimization domain defining
$\delta_1$ is nonempty, the definition of $\delta_1$ implies
\[
\delta_1
\leq
C_{\mathrm{feas}}\rho_N.
\]
\end{newrevblock}

\noindent Situation (ii): We now consider the situation where the basic solution $\bx'$ is a feasible solution to $V^{\mathrm{LP}}$ but not optimal. \begin{newrevblock}
Similarly, it holds that
\[
\bm r^\top\bm x'
\geq
\bm r^\top\widehat{\bm x}
-
C_{\mathrm{obj}}\rho_N,
\]
where
\[
C_{\mathrm{obj}}
:=
\|\bm r\|_\infty C_{\mathrm{bas}}.
\]
By the optimality of $\widehat{\bm x}$ for the empirical LP
and Claim~\ref{claim:Bound1},
\[
\bm r^\top\bm x'
\geq
V^{\mathrm{LP}}
-
\left(
C_{\mathrm{val}}+C_{\mathrm{obj}}
\right)\rho_N.
\]
If $\bm x'$ is feasible but suboptimal and the minimization
domain defining $\delta_2$ is nonempty, then
\[
\delta_2
\leq
\left(
C_{\mathrm{val}}+C_{\mathrm{obj}}
\right)\rho_N.
\]
\end{newrevblock}
\begin{newrevblock}
Define
\[
C_{\mathrm{id}}
:=
\max\left\{
C_{\mathrm{feas}},\,
C_{\mathrm{val}}+C_{\mathrm{obj}}
\right\}.
\]
Together with the definition
\[
\Delta_{\mathrm{id}}
=
\min\left\{
\eta_{\mathrm{id}},
\frac{\Delta}{2C_{\mathrm{id}}}
\right\},
\]
suppose that $\rho_N\leq\Delta_{\mathrm{id}}$.

If $\bm x'$ were infeasible and $\delta_1<\infty$, then
\[
\delta_1
\leq
C_{\mathrm{feas}}\rho_N
\leq
C_{\mathrm{id}}
\frac{\Delta}{2C_{\mathrm{id}}}
=
\frac{\Delta}{2}
\leq
\frac{\delta_1}{2},
\]
which is impossible. If $\delta_1=+\infty$, every population
basic solution is feasible, so this case is excluded directly.

Similarly, if $\bm x'$ were feasible but suboptimal and
$\delta_2<\infty$, then
\[
\delta_2
\leq
\left(
C_{\mathrm{val}}+C_{\mathrm{obj}}
\right)\rho_N
\leq
\frac{\Delta}{2}
\leq
\frac{\delta_2}{2},
\]
which is impossible. If $\delta_2=+\infty$, every feasible
population basic solution is optimal.

Thus, $\bm x'$ is feasible and optimal for the population LP.
Since Assumption~\ref{assump:degeneracy} gives a unique optimal
basis,
\[
\widehat{\mathcal I}=\mathcal I^*,
\qquad
\widehat{\mathcal J}=\mathcal J^*.
\]
\end{newrevblock}
Our proof is thus completed.


\begin{newrevblock}
\subsection{Proof of Claim~\ref{claim:Bound1}}

Let
$(\widehat{\mathcal I},\widehat{\mathcal J})$
be the optimal basis selected by
Algorithm~\ref{alg:Idenbasis}, and define
\[
B
:=
A_{\widehat{\mathcal J},
  \widehat{\mathcal I}},
\qquad
\widehat B
:=
\widehat A_{\widehat{\mathcal J},
             \widehat{\mathcal I}}(\mathcal H).
\]
By Assumption~\ref{assump:degeneracy}, $B$ is nonsingular.
Moreover, on the event $\mathcal E$,
\[
\|\widehat B^\top-B^\top\|_1
=
\|\widehat B-B\|_\infty
\leq
d_1\rho_N.
\]
Since $\rho_N\leq\eta_{\mathrm{id}}$,
\[
\left\|
B^{-\top}
\left(
\widehat B^\top-B^\top
\right)
\right\|_1
\leq
M_{\mathrm{bas}}d_1\rho_N
\leq
\frac12.
\]
The standard inverse perturbation bound therefore gives
\[
\|\widehat B^{-\top}\|_1
\leq
2M_{\mathrm{bas}}.
\label{eq:empirical_dual_inverse}
\]

Let $\bm y^*$ be the optimal dual basic solution associated
with the population optimal basis, and let
$\widehat{\bm y}$ be the optimal dual basic solution
associated with
$(\widehat{\mathcal I},\widehat{\mathcal J})$.
They satisfy
\[
\bm y^*_{(\mathcal J^*)^c}
=
\bm0,
\qquad
A_{\mathcal J^*,\mathcal I^*}^{\top}
\bm y^*_{\mathcal J^*}
=
\bm r_{\mathcal I^*},
\]
and
\[
\widehat{\bm y}_{\widehat{\mathcal J}^c}
=
\bm0,
\qquad
\widehat B^\top
\widehat{\bm y}_{\widehat{\mathcal J}}
=
\bm r_{\widehat{\mathcal I}}.
\]
Since both vectors are dual feasible,
\[
\bm y^*\geq\bm0,
\qquad
\widehat{\bm y}\geq\bm0.
\]
Consequently,
\[
\|\bm y^*\|_1
\leq
M_{\mathrm{bas}}\|\bm r\|_1,
\qquad
\|\widehat{\bm y}\|_1
\leq
2M_{\mathrm{bas}}\|\bm r\|_1.
\label{eq:dual_norm_bounds}
\]

Let $\widehat{\bm x}$ denote the empirical optimal basic
solution selected by Algorithm~\ref{alg:Idenbasis}. Since
$\widehat{\bm x}$ is feasible for the empirical LP,
\[
\widehat A(\mathcal H)\widehat{\bm x}
\leq
\bm c.
\]
On the event $\mathcal E$,
\[
A\widehat{\bm x}
\leq
\widehat A(\mathcal H)\widehat{\bm x}
+
\rho_N\|\widehat{\bm x}\|_1\bm1
\leq
\bm c+B_x\rho_N\bm1.
\]
Using dual feasibility of $\bm y^*$ and strong duality,
\begin{equation}
\begin{aligned}
\widehat V^{\mathrm{LP}}(\mathcal H)
&=
\bm r^\top\widehat{\bm x} \\
&=
(\bm y^*)^\top A\widehat{\bm x} \\
&\leq
(\bm y^*)^\top\bm c
+
B_x\|\bm y^*\|_1\rho_N \\
&\leq
V^{\mathrm{LP}}
+
M_{\mathrm{bas}}B_x
\|\bm r\|_1\rho_N.
\end{aligned}
\label{eq:value_upper}
\end{equation}

Applying Assumption~\ref{assump:degeneracy} at
$\widetilde A=A$ gives
\[
\|\bm x^*\|_1\leq B_x.
\]
Since $\bm x^*$ is feasible for the population LP,
\[
\widehat A(\mathcal H)\bm x^*
\leq
A\bm x^*
+
\rho_N\|\bm x^*\|_1\bm1
\leq
\bm c+B_x\rho_N\bm1.
\]
Using dual feasibility of $\widehat{\bm y}$ and strong
duality for the empirical LP,
\begin{equation}
\begin{aligned}
V^{\mathrm{LP}}
&=
\bm r^\top\bm x^* \\
&=
\widehat{\bm y}^{\top}
\widehat A(\mathcal H)\bm x^* \\
&\leq
\widehat{\bm y}^{\top}\bm c
+
B_x\|\widehat{\bm y}\|_1\rho_N \\
&\leq
\widehat V^{\mathrm{LP}}(\mathcal H)
+
2M_{\mathrm{bas}}B_x
\|\bm r\|_1\rho_N.
\end{aligned}
\label{eq:value_lower}
\end{equation}

Combining~\eqref{eq:value_upper}
and~\eqref{eq:value_lower} yields
\[
\left|
V^{\mathrm{LP}}
-
\widehat V^{\mathrm{LP}}(\mathcal H)
\right|
\leq
2M_{\mathrm{bas}}B_x
\|\bm r\|_1\rho_N.
\]
This proves the claim.
\end{newrevblock}

\section{Proof of \Cref{thm:Robust}}
We now complete the proof of \Cref{thm:Robust}. Note that the output $\bx$ of \Cref{alg:Idenbasis} is an optimal solution to $\hat{V}^{\mathrm{LP}}(\mathcal{H})$. We only need to bound the gap between $V^{\mathrm{LP}}$ and $\hat{V}^{\mathrm{LP}}(\mathcal{H})$. We now condition on the event that 
\begin{equation}\label{def:event0}
\begin{aligned}
\mathcal{E}=\Biggl\{\left|\widehat A_{k,i}(\mathcal H)-A_{k,i}\right|
\leq \Rad(N, \blue{\eps_0}), ~~\forall k\in[K], ~\forall i\in[d_1]  \Biggr\}.
\end{aligned}
\end{equation}
We know that this event $\mathcal{E}$ happens with probability at least \blue{$1-Kd_1\eps_0=1-\eps$}. 

Note that since $\bx$ is a feasible solution to $\hat{V}^{\mathrm{LP}}(\mathcal{H})$, we must have that
\[
\hat{A}(\mathcal{H})\bx\leq\bc
\]
which implies that
\begin{newrevblock}
\[
\begin{aligned}
A\bm x
&\leq
\widehat A(\mathcal H)\bm x
+
\left(A-\widehat A(\mathcal H)\right)\bm x\\
&\leq
\bm c
+
\Rad(N,\epsilon_0)\|\bm x\|_1\bm 1\\
&\leq
\bm c
+
B_x\Rad(N,\epsilon_0)\bm 1,
\end{aligned}
\]
where the final inequality follows from
the local boundedness clause of Assumption~\ref{assump:degeneracy}.
\end{newrevblock}

On the other hand, from \Cref{claim:Bound1},  we have that 
\[
|V^{\mathrm{LP}}-\bm r^\top\bm x|\leq \blue{C_{\mathrm{val}} \operatorname{Rad}(N,\epsilon_0)}.
\]
which completes our proof.

\section{Relating the Objective Value to the Terminal Residual}\label{pf:Lemma3}

\begin{lemma}\label{lem:RemainRe}
Denote by $\mathcal{I}^*$ and $\mathcal{J}^*$ the optimal basis identified by \Cref{alg:Idenbasis}. We also denote by $\bm{x}^*$ the corresponding optimal solution and $\by^*$ the corresponding optimal dual solution. Then, it holds that
\begin{equation}\label{eqn:Redompose}
N\cdot V^{\mathrm{LP}}-\sum_{n=1}^N \bm{r}^\top\mathbb{E}[\bm{x}^n]\leq \sum_{(s,a)\in\mathcal{J}^*} y^*_{(s,a)}\cdot \mathbb{E}[\bc^{\blue{N+1}}_{(s,a)}].
\end{equation}
\end{lemma}

From the update of \Cref{alg:Twophase}, we know that
\[\begin{aligned}
\sum_{n=1}^N (\bm{r})^\top\mathbb{E}[\bm{x}^n]= \sum_{n=1}^N (\bm{r}_{\mathcal{I}^*})^\top\mathbb{E}\left[\bm{x}^n_{\mathcal{I}^*}\right] 
\end{aligned}\]
Denote by $\bm{x}^*$ and $\bm{y}^*$ the optimal primal-dual variable corresponding to the optimal basis $\mathcal{I}^*$ and $\mathcal{J}^*$.
Since $(\mathcal{I}^*,\mathcal J^*)$ is an optimal basis, according to the optimal basis condition, we know that
\begin{equation}\label{eqn:020204}
A_{\mathcal{J}^*, \mathcal{I}^*}^\top\bm{y}^*_{\mathcal{J}^*}=\bm{r}_{\mathcal{I}^*}.
\end{equation}
\blue{Let $\mathcal F_{\mathrm I}$ be the sigma-field generated by the data used in \Cref{alg:Idenbasis} and the identified basis, and define $\mathcal G_n:=\sigma(\mathcal F_{\mathrm I},\mathcal H^n,\bc^n)$. Then $\bx^n$ is $\mathcal G_n$-measurable and $\mathbb E[A^n\mid\mathcal G_n]=A_{\mathcal J^*,\mathcal I^*}$; hence, by the tower property,}
\begin{equation}\label{eqn:011801}
\begin{aligned}
\sum_{n=1}^N (\bm{r}_{\mathcal{I}^*})^\top\mathbb{E}\left[\bm{x}^n_{\mathcal{I}^*}\right]&=\sum_{n=1}^N \left( (A_{\mathcal{J}^*, \mathcal{I}^*}^\top\bm{y}_{\mathcal{J}^*}^*\right)^\top \mathbb{E}[\bm{x}^n_{\mathcal{I}^*}]\\
&\blue{=\mathbb{E}\left[ \sum_{n=1}^N \left( (A^n)^\top\bm{y}_{\mathcal{J}^*}^* \right)^\top\bm{x}^n_{\mathcal{I}^*} \right]}\\
&=\mathbb{E}\left[ \sum_{n=1}^N (\bm{y}^*)^\top A^n \bm{x}^n_{\mathcal{I}^*} \right]
\end{aligned}
\end{equation}
From the update rule \eqref{eqn:UpdateAlpha2}, we have
\begin{equation}\label{eqn:011802}
\sum_{n=1}^N A^n\bm{x}^n_{\mathcal{I}^*}=\bm{c}^{1}_{\mathcal{J}^*}-\bm{c}^{\blue{N+1}}_{\mathcal{J}^*}.
\end{equation}
Plugging \eqref{eqn:011802} back into \eqref{eqn:011801}, we get that
\[
\sum_{n=1}^N (\bm{r}_{\mathcal{I}^*})^\top\mathbb{E}\left[\bm{x}^n_{\mathcal{I}^*}\right]=(\bm{y}_{\mathcal{J}^*}^*)^\top \bm{c}^{1}_{\mathcal{J}^*} - (\bm{y}_{\mathcal{J}^*}^*)^\top \mathbb{E}\left[\bm{c}^{\blue{N+1}}_{\mathcal{J}^*} \right].
\]
Note that from the strong duality of $V^{\mathrm{LP}}$, we have
\[
N\cdot V^{\mathrm{LP}}=(\bm{y}_{\mathcal{J}^*}^*)^\top \bm{c}^1_{\mathcal{J}^*}.
\]
Then, we have that
\begin{equation}\label{eqn:020201}
\begin{aligned}
&N\cdot V^{\mathrm{LP}}-\sum_{n=1}^N (\bm{r})^\top\mathbb{E}[\bm{x}^n]
\leq (\bm{y}_{\mathcal{J}^*}^*)^\top\mathbb{E}\left[\bm{c}^{\blue{N+1}}_{\mathcal{J}^*}\right].
\end{aligned}
\end{equation}
Our proof is thus completed.

\section{Proof of \Cref{thm:BoundRe}}\label{pf:Thm2}
\begin{newrevblock}
Condition on the correct-basis event $\mathcal B_{\mathrm I}$ and set
$\epsilon_{\mathrm{II}}=1/(2N)$. Define
\[
\begin{aligned}
L_{\mathrm{II}}
&:=
\log\left(
\frac{2Nd_2^2}{\epsilon_{\mathrm{II}}}
\right),\\
r_n
&:=
\sqrt{\frac{L_{\mathrm{II}}}{2n}}
=
\Rad\left(
n,
\frac{\epsilon_{\mathrm{II}}}{Nd_2^2}
\right),
\qquad n\in[N].
\end{aligned}
\]
\end{newrevblock}
We consider the stochastic process $\tilde{\bc}_{(s,a)}(n)$ defined in \eqref{eqn:Average}. For a fixed $\nu>0$ which we specify later, we define a set
\begin{equation}\label{eqn:defX}
\mathcal{C}=\{ \bm{\bc}'\in\mathbb{R}^{|\mathcal{J}^*|}: c'_{(s,a)}\in[c_{(s,a)}-\nu, c_{(s,a)}+\nu], \forall (s,a)\in\mathcal{J}^* \}.
\end{equation}
It is easy to see that initially, $\tilde{\bm{c}}_{\mathcal{J}^*}(1)\in\mathcal{C}$. We show that $\tilde{\bm{c}}_{\mathcal{J}^*}(n)$ behaves well as long as they stay in the region $\mathcal{C}$ for a
sufficiently long time. 
To this end, we define a stopping time
\begin{equation}\label{eqn:Stoptime2}
\blue{\tau = \min\big(
\left\{n\in[N]:\widetilde{\mathbf c}_{\mathcal J^*}(n)\notin\mathcal C\right\}
\cup{\{N+1\}}
\big)}.
\end{equation}
Under this convention, 
\(\widetilde{\mathbf c}_{\mathcal J^*}(n)\in\mathcal C\) 
holds for all \(n \blue{<} \tau\). If the process never leaves
\(\mathcal C\) over the horizon, then \(\tau=\blue{N+1}\).
Note that in \Cref{alg:Twophase}, to prevent $\bm{x}^n$ from behaving ill when $n$ is small, we project it to a set that guarantees $\|\bm{x}^n\|_1\leq C$. We now show in the following lemma that when $n$ is large enough but smaller than the stopping time $\tau$, it is automatically satisfied that $\|\bx^n\|_1\leq C$.
\begin{lemma}\label{lem:projection}
\blue{On $\mathcal E_{\mathrm{II}}$, if $N_0'\leq n\blue{<}\tau$ and $\nu\leq\nu_0$, then $\hat A_{\mathcal J^*,\mathcal I^*}(\mathcal H^n)$ is nonsingular and $\|\tilde{\bx}_{\mathcal I^*}^n\|_1\leq C$.}
\begin{equation}\label{eqn:N0prime}
\blue{N_0'
=
\left\lceil
\frac{8d_2^2L_{\mathrm{II}}}{\sigma^2}
\right\rceil.}
\end{equation}
Also, $\nu_0$ is given as follows
\begin{equation}\label{eqn:nu0}
\blue{\nu_0 = \frac{\sigma C}{8d_2}.}
\end{equation}
\end{lemma}
\begin{newrevblock}
For the remainder of Appendix~\ref{pf:Thm2}, set
\[
\nu=\nu_0=\frac{\sigma C}{8d_2}.
\]
Under Assumption~\ref{assump:FunctionApp} and
$\|\bm x^n\|_1\leq C$, we may take
\[
c_4:=1+2C.
\]
Indeed, for every $k\in\mathcal J^*$ and $n\in[N]$,
\[
\left|
(A^n_{k,:}-A_{k,\mathcal I^*})\bm x^n_{\mathcal I^*}
\right|
\leq c_4,
\qquad
\left|
c_k-A^n_{k,:}\bm x^n_{\mathcal I^*}
\right|
\leq c_4.
\]
\end{newrevblock}
We bound $\mathbb{E}[\blue{N+1-\tau}]$ in the following lemma.
\begin{lemma}\label{lem:Stoptime}
Let the stopping time $\tau$ be defined in \eqref{eqn:Stoptime2}. It holds that
\[
\mathbb{E}[\blue{N+1-\tau}]\leq N_0'+
d_2\cdot G\blue{+N\eps_{\mathrm{II}}}
\]
where $G=\frac{
2\exp\left(-\nu^2/(16c_4^2)\right)
}{
\left(1-\exp\left(-\nu^2/(16c_4^2)\right)\right)^2
}$ and $N_0'$ is given in \eqref{eqn:N0prime}, as long as
\begin{equation}\label{eqn:022104}
\blue{\begin{aligned}
N\geq N_0', \nu\leq\nu_0,
\frac{(N_0'-1)c_4}{N-N_0'+1}\leq\frac{\nu}{4},
\sqrt{\frac{2L_{\mathrm{II}}}{N}}\log(N)\leq\frac{\nu}{4d_2C}.
\end{aligned}}
\end{equation}
Also, for any $N'$ such that $N_0'\leq N'\leq N$, it holds that
\begin{equation}\label{eqn:HighProbtau}
\begin{aligned}
&P(\tau\leq N')\\
&\blue{\leq \eps_{\mathrm{II}}+}\frac{2d_2}{1 - \exp\left({-\nu^2 / 16c_4^2}\right)}\cdot \exp\left( -\frac{\nu^2\cdot(N-N'+1)}{16\cdot c_4^2} \right).
\end{aligned}
\end{equation}
\end{lemma}
From the definition of the stopping time $\tau$ in \eqref{eqn:Stoptime2}, we know that for each $(s,a)\in\mathcal{J}^*$, it holds
\[
\blue{c_{(s,a)}^{\tau-1}\in[(N-\tau+2)\cdot(c_{(s,a)}-\nu), (N-\tau+2)\cdot(c_{(s,a)}+\nu)]}
\]
Thus, we have that
\begin{equation}\label{eqn:011906}
\blue{|c_{(s,a)}^{N+1}|\leq |c_{(s,a)}^{\tau-1}|+ \left|\sum_{n=\tau-1}^N A^n_{(s,a), :}\cdot \bx^n_{\mathcal{I}^*}\right|}
\end{equation}
and thus, \blue{with $C_c:=2\|\bc_{\mathcal J^*}\|_\infty+\nu+c_4$,}
\begin{equation}\label{eqn:011907}
\begin{aligned}
\blue{\left|\mathbb{E}[c_{(s,a)}^{N+1}]\right|} &\blue{\leq C_c\left(1+\mathbb{E}[N+1-\tau]\right)}\\
&\blue{\leq C_c(N_0' + d_2G+N\eps_{\mathrm{II}}+1)}.
\end{aligned}
\end{equation}
\begin{newrevblock}
Here $G$ is defined in \Cref{lem:Stoptime} and is independent of $N$;
moreover, $N\epsilon_{\mathrm{II}}=1/2$ and
$L_{\mathrm{II}}=2\log(2Nd_2)$. Hence
\[
\left|
\mathbb E[c_k^{N+1}]
\right|
\leq R_N,
\qquad k\in\mathcal J^*.
\]
Hence
\[
\left\|
\mathbb E
\left[
\bm c_{\mathcal J^*}^{N+1}
\mid
\mathcal B_{\mathrm I}
\right]
\right\|_\infty
\leq
R_N.
\]
Lemma~\ref{lem:RemainRe} therefore gives the objective bound in
Theorem~\ref{thm:BoundRe}, while the capacity identity gives the binding-constraint bound. This completes the part of Theorem~\ref{thm:BoundRe} proved in this appendix. Notice that the basis-identification sample size $N_{\mathrm I}$ does not enter this conditional resolving-phase bound; its query cost and failure probability are incorporated in \Cref{thm:sample,cor:query}.
\end{newrevblock}


\subsection{Proof of Lemma \ref{lem:projection}}
\begin{newrevblock}
Let
\[
B:=A_{\mathcal J^*,\mathcal I^*},
\qquad
\widehat B_n
:=
\widehat A_{\mathcal J^*,\mathcal I^*}(\mathcal H^n),
\qquad
\widetilde{\bm c}^{\,n}
:=
\frac{\bm c_{\mathcal J^*}^n}{N-n+1}.
\]
The true and empirical basis systems satisfy
\begin{equation}\label{eqn:OptQ3}
B\bm x^*_{\mathcal I^*}
=
\bm c_{\mathcal J^*},
\end{equation}
and, whenever $\widehat B_n$ is nonsingular,
\begin{equation}\label{eqn:OptQ4}
\widehat B_n\widetilde{\bm x}_{\mathcal I^*}^n
=
\widetilde{\bm c}^{\,n}.
\end{equation}
For $n<\tau$,
\begin{equation}\label{eqn:020301}
\left\|
\widetilde{\bm c}^{\,n}-\bm c_{\mathcal J^*}
\right\|_\infty
\leq\nu,
\qquad
\left\|
\widetilde{\bm c}^{\,n}-\bm c_{\mathcal J^*}
\right\|_1
\leq d_2\nu.
\end{equation}

Define the simultaneous resolving-phase event
\begin{equation}\label{def:event2}
\mathcal E_{\mathrm{II}}
:=
\bigcap_{n=1}^{N}
\bigcap_{k\in\mathcal J^*}
\bigcap_{i\in\mathcal I^*}
\left\{
\left|
\widehat A_{k,i}(\mathcal H^n)-A_{k,i}
\right|
\leq r_n
\right\}.
\end{equation}
Conditional on $\mathcal B_{\mathrm I}$, Hoeffding's inequality and a
union bound over the $Nd_2^2$ entry--iteration pairs give
\[
\mathbb P(
\mathcal E_{\mathrm{II}}
\mid
\mathcal B_{\mathrm I}
)
\geq
1-\epsilon_{\mathrm{II}}.
\]

On $\mathcal E_{\mathrm{II}}$,
\[
\|\widehat B_n-B\|_1\leq d_2r_n.
\]
If $n\geq N_0'$, then the definition of $N_0'$ gives
$d_2r_n\leq\sigma/4$. Hence
\[
\|B^{-1}(\widehat B_n-B)\|_1
\leq\frac14,
\]
so $\widehat B_n$ is nonsingular and
\[
\|\widehat B_n^{-1}\|_1
\leq
\frac{\|B^{-1}\|_1}
{1-\|B^{-1}(\widehat B_n-B)\|_1}
\leq
\frac{2}{\sigma}.
\]
Since $\widehat B_n$ is nonsingular,
\[
\widehat B_n^\dagger=\widehat B_n^{-1}.
\]
Therefore, the iterate defined in Algorithm~\ref{alg:Twophase} satisfies
\[
\widehat B_n\widetilde{\bm x}_{\mathcal I^*}^n
=
\widetilde{\bm c}^{\,n}.
\]
Subtracting \eqref{eqn:OptQ3} from \eqref{eqn:OptQ4} gives
\[
\widehat B_n
\left(
\widetilde{\bm x}_{\mathcal I^*}^n
-
\bm x_{\mathcal I^*}^*
\right)
=
\left(
\widetilde{\bm c}^{\,n}
-
\bm c_{\mathcal J^*}
\right)
+
(B-\widehat B_n)\bm x_{\mathcal I^*}^*.
\]
Therefore, for $N_0'\leq n<\tau$,
\[
\left\|
\widetilde{\bm x}_{\mathcal I^*}^n
-
\bm x_{\mathcal I^*}^*
\right\|_1
\leq
\frac{2}{\sigma}
\left(
d_2\nu
+
d_2r_n\|\bm x_{\mathcal I^*}^*\|_1
\right).
\]
Using
$\nu=\sigma C/(8d_2)$,
$d_2r_n\leq\sigma/4$, and
$\|\bm x_{\mathcal I^*}^*\|_1\leq C/2$, the right-hand side is at most
$C/2$. Consequently,
\[
\|\widetilde{\bm x}_{\mathcal I^*}^n\|_1
\leq
\|\bm x_{\mathcal I^*}^*\|_1+\frac C2
\leq C.
\]
Thus the projection in Algorithm~\ref{alg:Twophase} is inactive and
$\bm x^n=\widetilde{\bm x}^n$ for
$N_0'\leq n<\tau$.
\end{newrevblock}

\subsection{Proof of Lemma \ref{lem:Stoptime}}
Now we fix a $(s,a)\in\mathcal{J}^*$. We specify a $\bar{N}_0=N_0'$. \blue{For every $m\leq\bar N_0$, the update rule gives}
\[
\blue{\left|\tilde c_{(s,a)}(m)-c_{(s,a)}\right|
\leq\frac{(m-1)c_4}{N-m+1}
\leq\frac{(\bar N_0-1)c_4}{N-\bar N_0+1}\leq\frac{\nu}{4},}
\]
\blue{where the last inequality follows from \eqref{eqn:022104}. Since the bound is uniform over $(s,a)\in\mathcal J^*$, $\tau>\bar N_0$. We define $\xi_{(s,a)}(n)=\tilde{c}_{(s,a)}(n+1)-\tilde{c}_{(s,a)}(n)$ for $n\in[N-1]$. For any $\bar N_0<N'\leq N$, we have}
\begin{equation} \nonumber
\begin{aligned}
&\blue{\tilde{c}_{(s,a)}(N')-c_{(s,a)}}\\
&\blue{=\tilde{c}_{(s,a)}(\bar N_0)-c_{(s,a)}
+\sum_{n=\bar{N}_0}^{N'-1}D_{(s,a)}(n)
+\sum_{n=\bar{N}_0}^{N'-1}\mathbb{E}[\xi_{(s,a)}(n)|\mathcal{G}_n],}
\end{aligned}
\end{equation}
\blue{where, by \eqref{eqn:Aveupalpha} and $\mathbb E[A^n\mid\mathcal G_n]=A_{\mathcal J^*,\mathcal I^*}$,}
\[
\blue{\begin{aligned}
\xi_{(s,a)}(n)&=\frac{\tilde{c}_{(s,a)}(n)-A^n_{(s,a),:}\bx_{\mathcal{I}^*}^n}{N-n},\\
D_{(s,a)}(n)&:=\xi_{(s,a)}(n)-\mathbb E[\xi_{(s,a)}(n)\mid\mathcal G_n]\\
&=-\frac{(A^n_{(s,a),:}-A_{(s,a),\mathcal I^*})\bx^n_{\mathcal I^*}}{N-n}.
\end{aligned}}
\]
Then, it holds that
\begin{equation}\label{eqn:011901}
\blue{|D_{(s,a)}(n)|\leq \frac{c_4}{N-n}.}
\end{equation}
\blue{Moreover, $D_{(s,a)}(n)$ is $\mathcal G_{n+1}$-measurable and $\mathbb E[D_{(s,a)}(n)\mid\mathcal G_n]=0$.}
We then recall the following martingale version of Hoeffding's
inequality, also known as the Azuma--Hoeffding inequality.
\begin{lemma}[Azuma--Hoeffding inequality]
Let $\{D_t,\mathcal{F}_t\}_{t=1}^T$ be a martingale difference sequence,
i.e.,
\[
    \mathbb{E}[D_t\mid \mathcal{F}_{t-1}]=0,\qquad t=1,\ldots,T.
\]
If there exist deterministic constants $a_t>0$ such that
$|D_t|\leq a_t$ almost surely for every $t$, then, for any $b>0$,
\[
    \mathbb{P}\left(
    \left|\sum_{t=1}^T D_t\right|\geq b
    \right)
    \leq
    2\exp\left(
    -\frac{b^2}{2\sum_{t=1}^T a_t^2}
    \right).
\]
\end{lemma}
\blue{For $\bar N_0<m\leq N'$, we have $\sum_{n=\bar N_0}^{m-1}(N-n)^{-2}\leq2/(N-m+1)$. Hence, the Azuma--Hoeffding inequality and a union bound give, for any $b>0$,}
\begin{equation}\label{eqn:011902}
\begin{aligned}
&\blue{P\left(\max_{\bar N_0<m\leq N'}\left|\sum_{n=\bar N_0}^{m-1}D_{(s,a)}(n)\right|\geq b\right)}\\
&\blue{\leq\sum_{m=\bar N_0+1}^{N'}2\exp\left(-\frac{b^2(N-m+1)}{4c_4^2}\right)}\\
&\blue{\leq \frac{2}{1 - \exp\left(-b^2/(4c_4^2)\right)}\exp\left(-\frac{b^2(N-N'+1)}{4c_4^2}\right).}
\end{aligned}
\end{equation}

\blue{On $\mathcal E_{\mathrm{II}}$, \Cref{lem:projection} gives $\bx^n=\tilde\bx^n$ and \eqref{eqn:OptQ4} for $N_0'\leq n\blue{<}\tau$.} Thus,
\[
\tilde{c}_{(s,a)}(n)=\hat{A}_{(s,a), \mathcal{I}^*}(\mathcal{H}^n)\cdot \bx^n_{\mathcal{I}^*}.
\]
It holds that
\begin{equation}\label{eqn:011903}
\begin{aligned}
\blue{\mathbb{E}[\xi_{(s,a)}(n)|\mathcal{G}_n]}&\blue{=\frac{(\hat{A}_{(s,a), \mathcal{I}^*}(\mathcal{H}^n) - A_{(s,a), \mathcal{I}^*})\bx^n_{\mathcal I^*}}{N-n},}\\
\blue{\left|\mathbb{E}[\xi_{(s,a)}(n)|\mathcal{G}_n]\right|}&\blue{\leq\frac{d_2\cdot C\cdot r_n}{N-n}.}
\end{aligned}
\end{equation}
\blue{Consequently, under \eqref{eqn:022104}, for every $\bar N_0<m\leq N'\wedge\tau$,}
\begin{equation}\label{eqn:011904}
\begin{aligned}
&\blue{\frac{\sum_{n=\bar{N}_0}^{m-1}\left|\mathbb{E}[\xi_{(s,a)}(n)|\mathcal{G}_n]\right|}{d_2\cdot C}
\leq\sqrt{\frac{L_{\mathrm{II}}}{2}}\cdot\sum_{n=\bar{N}_0}^{m-1}\frac{1}{\sqrt{n}\cdot(N-n)}}\\
&\blue{\leq \sqrt{\frac{L_{\mathrm{II}}}{2}}\cdot \sqrt{m-1}\cdot \sum_{n=\bar{N}_0}^{m-1}\frac{1}{n\cdot(N-n)}}\\
&\blue{=\sqrt{\frac{L_{\mathrm{II}}}{2}}\cdot \frac{\sqrt{m-1}}{N}\cdot \sum_{n=\bar{N}_0}^{m-1}\left( \frac{1}{n}+\frac{1}{N-n} \right)}\\
&\blue{\leq \sqrt{2L_{\mathrm{II}}}\cdot \frac{\sqrt{m-1}}{N}\cdot \log(N)}\\
&\blue{\leq \sqrt{\frac{2L_{\mathrm{II}}}{N}}\cdot\log(N)}
\blue{\leq\frac{\nu}{4d_2C}},
\end{aligned}
\end{equation}

\blue{By the initial-displacement bound and \eqref{eqn:011904}, on $\mathcal E_{\mathrm{II}}$ the event $\{\tau\leq N'\}$ implies that, for some $(s,a)\in\mathcal J^*$, the maximum in \eqref{eqn:011902} is at least $\nu/2$. Applying \eqref{eqn:011902} with $b=\nu/2$ and adding $P(\mathcal E_{\mathrm{II}}^c)$ gives}
\begin{equation}\label{eqn:011905}
\begin{aligned}
&P(\tau\leq N') \\
&\blue{\leq \eps_{\mathrm{II}}+}\frac{2d_2}{1 - \exp\left({-\nu^2 / 16c_4^2}\right)}\cdot \exp\left( -\frac{\nu^2\cdot(N-N'+1)}{16\cdot c_4^2} \right).
\end{aligned}
\end{equation}
Therefore, we know that
\[
\begin{aligned}
\mathbb{E}[\blue{N+1-\tau}]&=\sum_{N'=1}^N P(\tau \leq N')\leq \bar{N}_0+\sum_{N'=\bar{N}_0}^N P(\tau \leq N')\\
&\leq \bar{N}_0+
d_2\cdot G\blue{+N\eps_{\mathrm{II}}} \nonumber
\end{aligned}
\]
\blue{This proves the lemma.}

\section{Completion of the Proof of \Cref{thm:BoundRe}
and Proof of \Cref{thm:sample}}
\begin{newrevblock}
Appendix~\ref{pf:Thm2} established the terminal-residual estimate and the objective bound in \Cref{thm:BoundRe}. We first derive the binding- and nonbinding-constraint bounds, thereby completing the proof of \Cref{thm:BoundRe}. We then remove the conditioning on the correct-basis event to prove \Cref{thm:sample}.

The convention that $\bar{\bm x}^N=\bm 0$ when the empirical LP does not return a finite optimal basic solution, together with the pseudoinverse and projection in Algorithm~\ref{alg:Twophase}, makes $\bar{\bm x}^N$ well defined on every sample path. First condition on the correct-basis event $\mathcal B_{\mathrm I}$ and write
\[
\bm x_{\mathrm{good}}^N
:=
\mathbb E[\bar{\bm x}^N\mid\mathcal B_{\mathrm I}].
\]
By the objective bound established in Appendix~\ref{pf:Thm2},
\[
V^{\mathrm{LP}}
-
\bm r^\top\bm x_{\mathrm{good}}^N
\leq
\frac{\|\bm y_{\mathcal J^*}^*\|_1R_N}{N}
\leq
\frac{\Lambda R_N}{N}.
\]
For $k\in\mathcal J^*$, the capacity update and conditional
unbiasedness give
\[
A_{k,:}\bm x_{\mathrm{good}}^N-c_k
=
-\frac{1}{N}
\mathbb E
\left[
 c_k^{N+1}
 \mid
 \mathcal B_{\mathrm I}
\right].
\]
Therefore,
\[
\left|
A_{k,:}\bm x_{\mathrm{good}}^N-c_k
\right|
\leq
\frac{R_N}{N}
\leq
\frac{\Lambda R_N}{N}.
\]

It remains to consider $k\in[K]\setminus\mathcal J^*$. Let
$B=A_{\mathcal J^*,\mathcal I^*}$. The capacity update and
\eqref{eqn:OptQ3} imply
\[
B\sum_{n=1}^N
\mathbb E[\bm x_{\mathcal I^*}^n\mid\mathcal B_{\mathrm I}]
=
N\bm c_{\mathcal J^*}
-
\mathbb E[
\bm c_{\mathcal J^*}^{N+1}
\mid\mathcal B_{\mathrm I}
],
\]
and hence
\[
\sum_{n=1}^N
\mathbb E[\bm x_{\mathcal I^*}^n\mid\mathcal B_{\mathrm I}]
=
N\bm x_{\mathcal I^*}^*
-
B^{-1}
\mathbb E[
\bm c_{\mathcal J^*}^{N+1}
\mid\mathcal B_{\mathrm I}
].
\]
Because $\bm x_{\mathcal I^{*c}}^n=0$ and $\bm x^*$ is feasible,
\[
\begin{aligned}
A_{k,:}\bm x_{\mathrm{good}}^N-c_k
&\leq
\frac{1}{N}
\left\|
A_{k,\mathcal I^*}B^{-1}
\right\|_1
\left\|
\mathbb E[
\bm c_{\mathcal J^*}^{N+1}
\mid\mathcal B_{\mathrm I}
]
\right\|_\infty\\
&\leq
\frac{\Lambda R_N}{N}.
\end{aligned}
\]
Thus, conditional on $\mathcal B_{\mathrm I}$,
\[
A\bm x_{\mathrm{good}}^N-\bm c
\leq
\frac{\Lambda R_N}{N}\bm 1.
\]

Together with the objective bound established in Appendix~\ref{pf:Thm2}, this completes the proof of \Cref{thm:BoundRe}. We now remove the conditioning. Let
$p:=\mathbb P(\mathcal B_{\mathrm I}^c)\leq\delta_{\mathrm I}$ and,
when $p>0$, define
\[
\bm x_{\mathrm{bad}}^N
:=
\mathbb E[
\bar{\bm x}^N
\mid
\mathcal B_{\mathrm I}^c
].
\]
The projection in Algorithm~\ref{alg:Twophase} gives
$\|\bar{\bm x}^N\|_1\leq C$ almost surely, and hence
$\|\bm x_{\mathrm{bad}}^N\|_1\leq C$. By
Assumption~\ref{assump:FunctionApp},
\[
V^{\mathrm{LP}}
-
\bm r^\top\bm x_{\mathrm{bad}}^N
\leq
C_{\mathrm{fail}},
\qquad
A\bm x_{\mathrm{bad}}^N-\bm c
\leq
C_{\mathrm{fail}}\bm 1.
\]
Since
\[
\bm x_{\mathrm{mean}}^N
=
(1-p)\bm x_{\mathrm{good}}^N
+
p\bm x_{\mathrm{bad}}^N,
\]
the law of total expectation gives
\[
V^{\mathrm{LP}}
-
\bm r^\top\bm x_{\mathrm{mean}}^N
\leq
\frac{\Lambda R_N}{N}
+
\delta_{\mathrm I}C_{\mathrm{fail}},
\]
and
\[
A\bm x_{\mathrm{mean}}^N-\bm c
\leq
\left(
\frac{\Lambda R_N}{N}
+
\delta_{\mathrm I}C_{\mathrm{fail}}
\right)\bm 1.
\]
By the definition
\[
\epsilon_N^{\mathrm{mean}}
=
\epsilon_N^{\mathrm{res}}
+
\delta_{\mathrm I}C_{\mathrm{fail}},
\]
the preceding two inequalities prove
Theorem~\ref{thm:sample}.

\end{newrevblock}

\section{Proof of \Cref{cor:value_function_policy}} \label{app:J}
\begin{newrevblock}
Let $T$ denote the optimal Bellman operator for the cost-minimization
problem,
\[
(TV)(s)
=
\min_{a\in\mathcal A}
\left\{
c(s,a)
+
\gamma
\mathbb E_{s'\sim P(\cdot\mid s,a)}[V(s')]
\right\},
\]
and let $T_\pi$ denote the Bellman operator induced by policy $\pi$.

Because the guarantee in \Cref{thm:sample} is assumed to hold for all
constraints of the full ALP,
\[
V_{\mathrm{mean}}^N(s)
-
\gamma
\mathbb E_{s'\sim P(\cdot\mid s,a)}
[V_{\mathrm{mean}}^N(s')]
\leq
c(s,a)+\epsilon,
\qquad
\forall(s,a).
\]
Equivalently,
\[
V_{\mathrm{mean}}^N
\leq
TV_{\mathrm{mean}}^N+\epsilon\bm 1.
\]
Define
\[
\beta:=\frac{\epsilon}{1-\gamma},
\qquad
\widetilde V_{\mathrm{mean}}^N
:=
V_{\mathrm{mean}}^N-\beta\bm 1.
\]
Then
\[
\widetilde V_{\mathrm{mean}}^N
\leq
T\widetilde V_{\mathrm{mean}}^N.
\]
By monotonicity and contraction of $T$,
\[
\widetilde V_{\mathrm{mean}}^N
\leq
V^{\pi^*}.
\]
Let
\[
e_N
:=
V^{\pi^*}-\widetilde V_{\mathrm{mean}}^N
\geq0.
\]

Using exact realizability, exact LP representation, and
$\mu(\mathcal S)=1$, the objective guarantee gives
\[
\begin{aligned}
\int_{\mathcal S}e_N(s)\,\mu(ds)
&=
V^{\mathrm{LP}}
-
\bm r^\top\bm x_{\mathrm{mean}}^N
+
\beta\\
&\leq
\epsilon+\frac{\epsilon}{1-\gamma}\\
&=
\frac{2-\gamma}{1-\gamma}\epsilon.
\end{aligned}
\]

A constant shift does not change the greedy policy, so the policy
$\pi_{\mathrm{mean}}^N$ in \eqref{eq:policy_mean} is also greedy with
respect to $\widetilde V_{\mathrm{mean}}^N$. Hence
\[
T_{\pi_{\mathrm{mean}}^N}
\widetilde V_{\mathrm{mean}}^N
=
T\widetilde V_{\mathrm{mean}}^N
\leq
V^{\pi^*}.
\]
It follows that
\[
T_{\pi_{\mathrm{mean}}^N}
\widetilde V_{\mathrm{mean}}^N
-
\widetilde V_{\mathrm{mean}}^N
\leq
e_N.
\]
Using the Bellman equation for
$V^{\pi_{\mathrm{mean}}^N}$,
\[
V^{\pi_{\mathrm{mean}}^N}
-
V^{\pi^*}
\leq
(I-\gamma P_{\pi_{\mathrm{mean}}^N})^{-1}e_N.
\]

For an initial distribution $\nu$ and a policy $\pi$, define the
discounted state-visitation distribution
\[
d_\nu^\pi
:=
(1-\gamma)
\sum_{t=0}^{\infty}
\gamma^t\nu P_\pi^t,
\]
and the concentrability coefficient
\[
C_{\nu,\mu}^{\pi}
:=
\left\|
\frac{d d_\nu^\pi}{d\mu}
\right\|_\infty.
\]
Integrating the preceding inequality with respect to $\nu$ yields
\[
\begin{aligned}
V^{\pi_{\mathrm{mean}}^N}(\nu)
-
V^{\pi^*}(\nu)
&\leq
\frac{1}{1-\gamma}
\int_{\mathcal S}
e_N(s)\,
d_\nu^{\pi_{\mathrm{mean}}^N}(ds)\\
&\leq
\frac{C_{\nu,\mu}^{\pi_{\mathrm{mean}}^N}}{1-\gamma}
\int_{\mathcal S}e_N(s)\,\mu(ds)\\
&\leq
\frac{(2-\gamma)
C_{\nu,\mu}^{\pi_{\mathrm{mean}}^N}}
{(1-\gamma)^2}
\epsilon.
\end{aligned}
\]
This proves the corollary.
\end{newrevblock}

\end{APPENDICES}

\end{document}